%% file: acl_latex.tex
\pdfoutput=1
\documentclass[11pt]{article}

\usepackage[preprint]{acl}
\usepackage{times}
\usepackage{latexsym}

\usepackage{booktabs}
\usepackage{multicol}
\usepackage{multirow}

\usepackage[T1]{fontenc}
\usepackage[utf8]{inputenc}

\usepackage{microtype}

\usepackage{inconsolata}

\usepackage{graphicx}
\usepackage{xspace}
\usepackage{amssymb}
\usepackage{amsmath}
\usepackage{enumitem}
\usepackage{xcolor}
\usepackage[capitalize]{cleveref}
\usepackage{listings}
\usepackage{hyperref}
\usepackage{pifont}
\usepackage{float}

\usepackage{multirow}
\usepackage{longtable}
\usepackage{listings}
\usepackage{graphicx}
\usepackage{mathtools}
\usepackage{amsthm}
\usepackage{xspace}
\usepackage[subtle]{savetrees}
\usepackage[most]{tcolorbox}% it depends on "graphicx" package
\usepackage{colortbl}
\usepackage{bbm}
\tcbuselibrary{breakable}

\newcommand{\ours}{\textsc{GEMS}\xspace}

\definecolor{berkeleyblue}{HTML}{3B7EA1}
\definecolor{berkeleygold}{HTML}{FDB515}
\definecolor{lightcyan}{HTML}{E5F2F2}
\definecolor{main}{HTML}{4472C4}    % setting main color to be used
\definecolor{sub}{HTML}{EBF4FF}     % setting sub color to be used

\definecolor{lightcyan}{rgb}{0.88,1,1}  % example definition

\newcommand{\verytiny}{\fontsize{5}{5}\selectfont}

\DeclareMathSizes{10}{9}{6}{5}

\title{Graph-Based Alternatives to LLMs for Human Simulation}

\author{
 \textbf{Joseph Suh},
 \textbf{Suhong Moon},
 \textbf{Serina Chang}
\\[0.5ex]
 University of California, Berkeley
\\
 \small{
   \{josephsuh,serinac\}@berkeley.edu
 }
}

\begin{document}
\maketitle

\begin{abstract}

\vspace{-8pt}
Large language models (LLMs) have become a popular approach for simulating human behaviors, yet it remains unclear if LLMs are necessary for all simulation tasks. 
We study a broad family of close-ended simulation tasks, with applications from survey prediction to test-taking, and show that a graph neural network
can match or surpass strong LLM-based methods. We introduce \textbf{G}raph-bas\textbf{E}d \textbf{M}odels for Human \textbf{S}imulation (\ours) which formulates close-ended simulation as link prediction on a heterogeneous graph of individuals and choices.
Across three datasets and three evaluation settings, \ours matches or outperforms the strongest LLM-based methods while using three orders of magnitude fewer parameters. These results suggest that graph-based modeling can complement LLMs as an efficient and transparent approach to simulating human behaviors.
Code is available at \url{https://github.com/schang-lab/gems}.
\end{abstract}

\input{main_sections/intro}
\input{main_sections/related_work}
\input{main_sections/problem}
\input{main_sections/methods} 
\input{main_sections/experiments}
\input{main_sections/advantages}
\input{main_sections/conclusion}
\input{appendix_sections/limitations}

\section*{Acknowledgments}

The authors thank Minwoo Kang, Prof. John Canny, and Prof. Emma Pierson for constructive comments and valuable feedback.
This work was supported in part by the Korea Foundation for Advanced Studies and Google Research Scholar Program, and by compute resources from VESSL AI, the Center for Human-Compatible AI at Berkeley, and the BAIR-Google Commons program.
The views and findings expressed are those of the authors and should not be interpreted as representing the official views or policies of any sponsor.

\bibliography{references}

\appendix

\input{appendix_sections/additional_related_work}
\input{appendix_sections/question_examples}
\input{appendix_sections/graph_statistics}
\input{appendix_sections/training_detail}
\input{appendix_sections/addtional_experiment}
\input{appendix_sections/ablations}
\input{appendix_sections/prompts}
\input{appendix_sections/note_gpt_oss}

\end{document}

%% file: main_sections/intro.tex
\section{Introduction}
\label{main:introduction}

\vspace{-6pt}
Human simulation has recently attracted significant attention,
driving new research directions \citep{gao2024agent}, workshops \citep{social-sims-llms-workshop,llm-persona-modeling-workshop},
panels \citep{hwang2025human}, and even startups \citep{expected-parrot, yc-artificial-societies}.
Throughout this excitement, large language models (LLMs) have remained by far the predominant approach,
to the extent that references to this burgeoning field typically include LLM in their titles, such as ``LLM social simulation'' \citep{anthis2025llm} or ``LLM-simulated data'' \citep{hwang2025human}.
While some simulation tasks are open-ended \citep{zhousotopia,bianchi2024well},
many of the most popular tasks are \textit{close-ended}, predicting an individual's response from a set of options.
Despite this close-ended structure, LLMs have also remained the predominant approach here,
with many works proposing LLM-based methods for problems such as predicting survey responses
\citep{santurkar2023whose,hwang2023aligning,zhaogroup,feng-etal-2024-modular,moon2024virtual,suh2025language,cao2025specializing,kolluri2025finetuningllmshumanbehavior,krsteski2025valid},
effects of social science experiments \citep{hewitt2024predicting,park2024generative,manning2024automated},
voting outcomes \citep{yu2024large,kreutner2025persona,von2025vox,li2025llm},
and test-taking abilities \citep{wang2025adaptive,binz2025foundation}.

\vspace{-4pt}
While LLMs have shown strong results on these tasks,
especially after fine-tuning \citep{suh2025language,cao2025specializing,binz2025foundation},
LLMs have downsides: they are expensive to run and train and their opaque pretraining processes lead to concerns of data leakage \citep{deng2024unveiling} and social biases \citep{cheng2023compost,bisbee2024synthetic}. 
This motivates a natural question:
\textbf{for close-ended simulation tasks, can a smaller and more transparent model class be competitive with LLMs?} 
On one hand, focusing on these popular close-ended tasks may reduce LLMs' comparative advantage in open-ended text generation. 
On the other hand, high-quality simulation may still depend on unique LLM capabilities---language understanding, knowledge from pretraining, and effective adaptation via prompting or fine-tuning. 
Resolving this tension is important both for understanding what drives performance in close-ended simulation
and for identifying modeling approaches that may be more efficient or transparent without sacrificing predictive performance.

\begin{figure*}[t]
\centering
\includegraphics[width=0.9\textwidth]{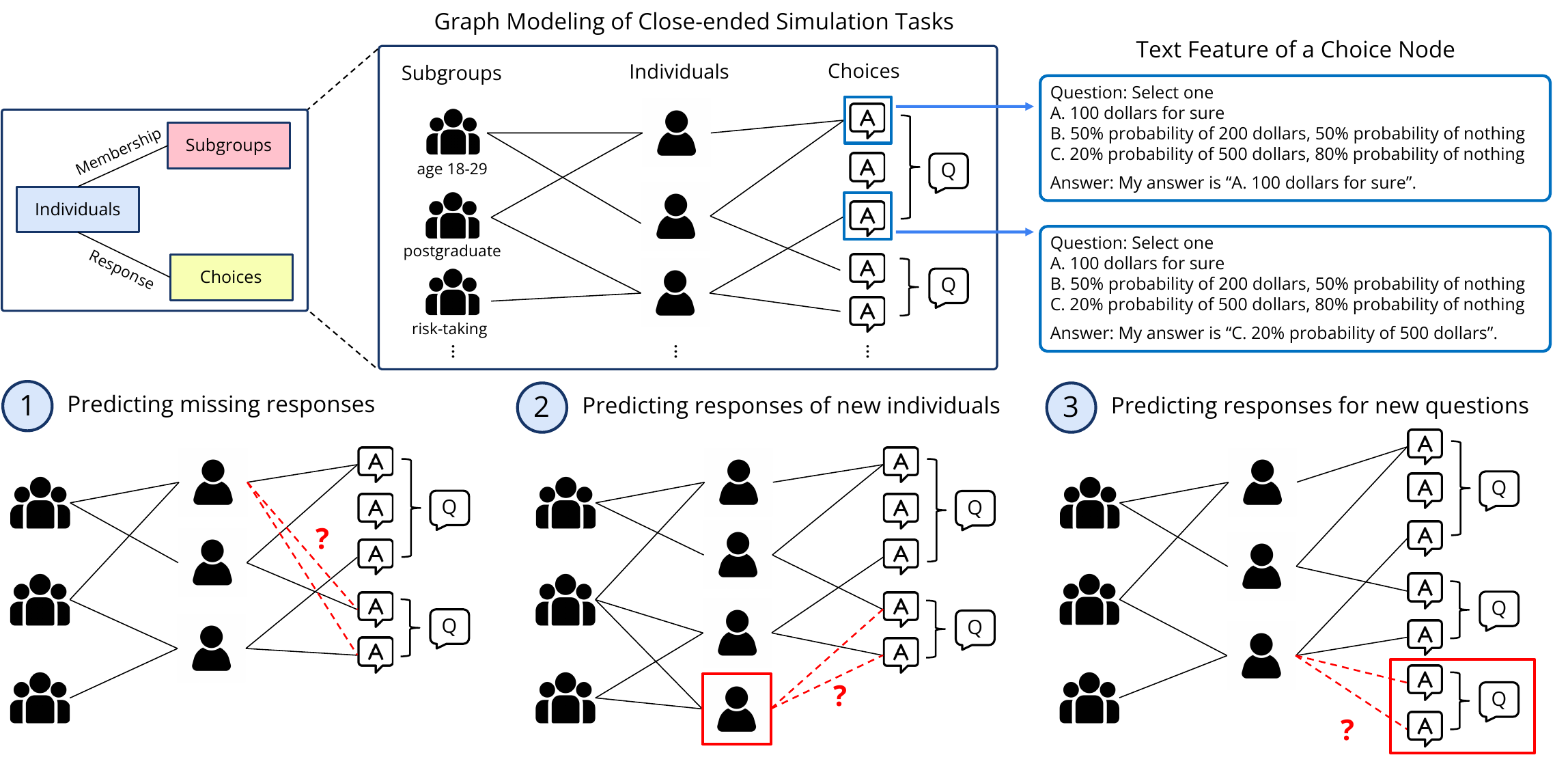}
\vspace{-5pt}
\caption{
In \ours, we construct a heterogeneous graph for close-ended simulation tasks (\textbf{Top})
where the goal is to predict the option chosen by an individual in response to a context.
Under three widely studied settings (\textbf{Bottom}),
we show that our approach achieves prediction accuracy consistently comparable to strong LLM-based methods.
}
\label{fig:figure_1}
\end{figure*}

\vspace{-4pt}
\textbf{The present work.}
We show that a simpler model---graph neural networks (GNNs)---can indeed match or outperform LLMs in close-ended simulation settings, with far less compute and greater transparency.
We introduce \textbf{G}raph-bas\textbf{E}d \textbf{M}odels for Human \textbf{S}imulation (\ours), which formulates close-ended simulation tasks as link prediction on a heterogeneous graph, with nodes as individuals and options and edges as observed choices (\Cref{fig:figure_1}). \ours uses a GNN to learn representations of individuals, subgroups, and choices and to predict unseen choices.
We consider three key settings for close-ended simulation tasks: predicting (1) missing responses (imputation), (2) responses of new individuals, and (3) responses for new questions. We compare \ours against strong LLM-based methods---zero-shot, few-shot, chain-of-thought prompting, and supervised fine tuning---on three datasets: \textsc{OpinionQA} public opinion polls \citep{santurkar2023whose}, \textsc{Twin-2K} psychological / economic / personality measures \citep{toubia2025twin}, and tests of individuals' grammar skills \citep{jansen2021rational, binz2025foundation}. 

Across the datasets and settings, \ours matches or outperforms strong LLM-based methods used in prior work.
In the first two settings, \ours achieves such performance using only the GNN and no language representations. In the third setting, where generalizing to unseen text is essential, leveraging a lightweight projection layer mapping frozen LLM representations to GNN embeddings \citep{sheng2025language} yields comparable performance.
While matching LLM accuracy, \ours uses $\sim10^3$ fewer parameters and up to $10^2$ times less compute, lowering barriers for researchers with limited compute and enabling scaling to larger datasets. 
Furthermore, unlike LLMs' opaque pretraining, \ours can be trained from scratch on transparent, domain-specific data, mitigating data leakage and bias concerns.
Overall, our work offers graph-based modeling as a compelling alternative to LLMs for human simulation, enabling broader participation in this interdisciplinary field and more trustworthy predictions.

%% file: main_sections/related_work.tex
\section{Related Work}
\label{main:related_work}

As interest in human simulation has risen sharply in the past few years, 
LLMs have remained by far the predominant approach \citep{gao2024agent,anthis2025llm,hwang2025human}.
As described previously, a large portion of this literature focuses on close-ended simulation tasks,
aiming to predict an individual's choice among a set of options, typically cast as next-token prediction of the LLM.
Prior work has explored prompting strategies \citep{dominguez2023questioning},
including few-shot prompting \citep{hwang2023aligning} and prompt engineering \citep{kim2025few},
as well as conditioning on open-ended narratives \citep{park2024generative, moon2024virtual, rahimzadeh2025synthia}.
Recently, fine-tuning has emerged as a promising alternative, either on community-specific text corpora \citep{chu2023language, he2024community, li2024culturellm, feng-etal-2024-modular}
or directly on human choice data \citep{cao2025specializing,suh2025language,binz2025foundation,xie2025fm,kolluri2025finetuningllmshumanbehavior},
augmented with prediction-powered inference \citep{angelopoulos2023prediction, krsteski2025valid}.

Yet the task remains selecting from a small, fixed set of options (or tokens), where LLM-based methods might not be the sole approach.
We build on this observation and, in \ours, emphasize the relational structure underlying human choices.
This approach exploits similar relational dependencies as graph-based recommender systems \citep{ying2018graph,fan2019graph,he2020lightgcn}, but have been absent in the LLM-centric simulation literature.
Our work is the first to clarify how LLM methods perform in comparison to graph-based models in human choice simulation and to provide a direct comparison.
In the Appendix (Section~\ref{appendix:additional_related_work}), we provide an extended discussion of graph-based recommenders, along with classical models (e.g., from economics) for modeling discrete choice.

%% file: main_sections/problem.tex
\section{Problem Definition}
\label{main:formulation}

We focus on close-ended simulation tasks,
where the goal is to predict an individual's selected response among a set of options.
Given an individual $u$ and a question $q$, with answer options $\mathcal{A}(q)$, the goal is to predict $u$'s response $y_{uq} \in \mathcal{A}(q)$.
Each individual has \textit{individual features}, such as demographic variables.
We use individual features to define \textit{subgroups}, which are groups of individuals sharing one or more features.
We also have \textit{question features} and \textit{option features}.
Since we focus on simulation tasks where LLM-based methods have been used,
these features are text, i.e., the text of the question and of each option,
but our framework is not restricted to text-only features.
We define a \textit{choice} as a pair $(q, a)$ of question $q$ and answer option $a \in \mathcal{A}(q)$;
its \textit{choice feature} is the concatenation of the question text and option text.
We observe a set of prior responses $\mathcal{Y}$,
which consists of responses from \textit{seen} individuals (i.e., those with at least one response in $\mathcal{Y}$) and \textit{seen} questions (i.e., those with at least one response in $\mathcal{Y}$).
However, we do not observe responses between all pairs of seen individuals and questions.

We consider three settings of simulating human choices widely studied in previous work (Figure~\ref{fig:figure_1}).

\vspace{-10pt}
\begin{description}[leftmargin=0pt,labelindent=0pt]
\item[\textbf{(1) Missing response (Imputation).}]
Given a \emph{seen} individual $u$ with individual features
and a \textit{seen} question $q$ with question and option features,
predict the missing response $y_{uq} \notin \mathcal{Y}$.
Prior work studies few-shot prompting and fine-tuning for this setting \citep{hwang2023aligning, zhaogroup, kim2025few, kolluri2025finetuningllmshumanbehavior}.

\vspace{-5pt}
\item[\textbf{(2) New individuals.}]
Given a \textit{new} individual $u$, where we observe their individual features but not any prior responses, predict $u$'s responses to seen questions.
This setting has been investigated in several simulation works \citep{santurkar2023whose, moon2024virtual, kang2025higher, li2025llm}
and is also of interest to pluralistic alignment \citep{feng-etal-2024-modular, yaono}.

\vspace{-5pt}
\item[\textbf{(3) New questions.}]
Given a \textit{new} question $q$, where we observe its question and option features but not any prior responses, predict the responses of seen individuals to $q$.
This setting is useful for simulating newly designed items in survey research \citep{rothschild2024opportunities, suh2025language, cao2025specializing} or testing generalization to new simulation settings \citep{binz2025foundation, xie2025fm}.
\end{description}

%% file: main_sections/methods.tex
\section{\ours Graph-based Modeling}
\label{main:architecture}

\subsection{Graph Representation of the Task}
\label{main:graph_formulation}

We represent the task as a heterogeneous graph $\mathcal{G}$ with three types of nodes: subgroups $\mathcal{S}$, individuals $\mathcal{U}$, and choices $\mathcal{C}$.
Choice nodes are structured as a disjoint union $\mathcal{C} = \mathcal{C}_1 \cup \mathcal{C}_2 \cup \cdots \cup \mathcal{C}_{n}$,
where $\mathcal{C}_q$ is the set of choice nodes for question $q$ and $n$ is the total number of questions.
We include two bidirectional relations: membership and response.
Membership edges $E_{\mathcal{US}}$ with an adjacency matrix $\smash{\mathbf{A}_{\mathcal{US}}\!\in\!\{0,1\}^{|\mathcal{U}|\times|\mathcal{S}|}}$ connect each individual to relevant subgroups.
Response edges $E_{\mathcal{UC}}$ with an adjacency matrix $\smash{\mathbf{A}_{\mathcal{UC}}\!\in\!\{0,1\}^{|\mathcal{U}|\times|\mathcal{C}|}}$ record which choice an individual made as a response to a question.
Because each question requires selecting one choice, the row-wise sum of $\mathbf{A}_{\mathcal{UC}}$ is at most $n$.

\begin{figure*}[t]
    \centering
    \includegraphics[width=0.9\textwidth]{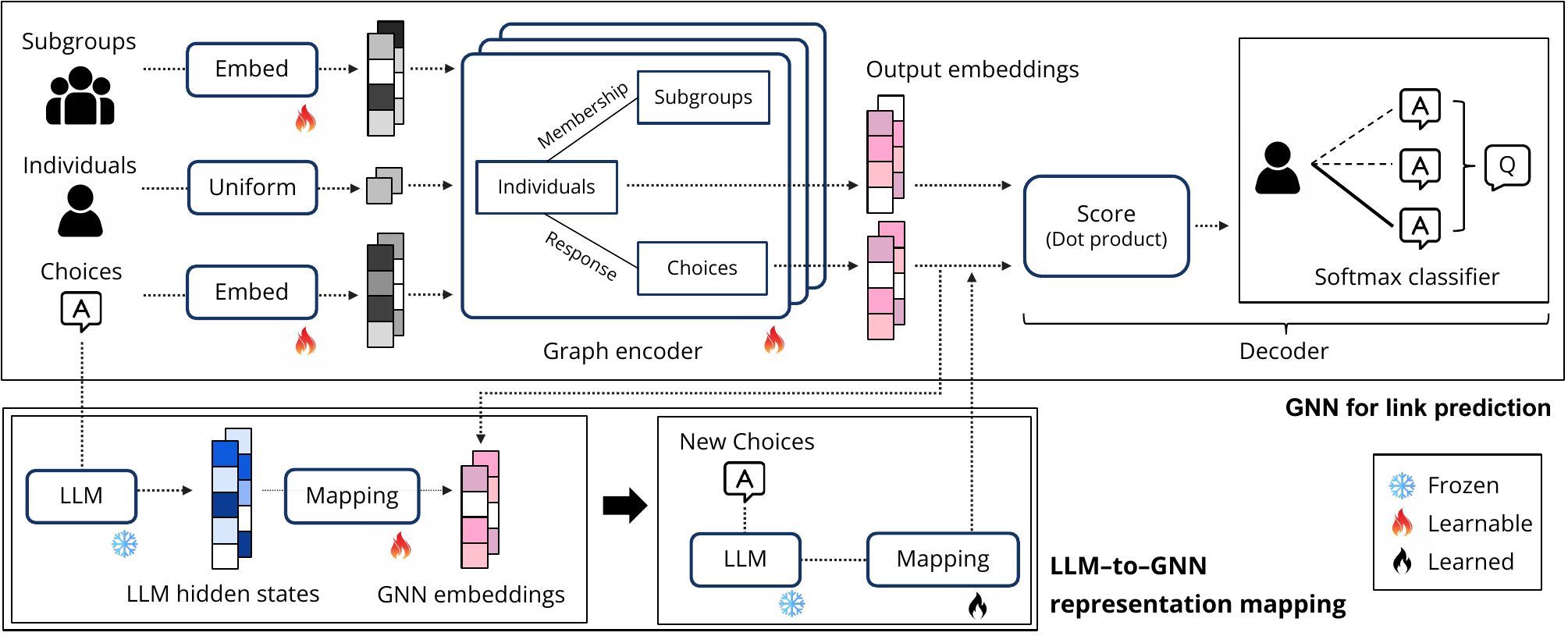}
    \caption{
    Overall architecture of \ours. The graph encoder learns representations of individual and choice nodes from the relational structure of observed responses, then predicts new responses with a softmax classifier over question options (\textbf{Top}).
    In setting 3, we train a simple LLM-to-GNN projection that maps a language representation of the choice node's text to its GNN embedding, so that we can acquire meaningful representations of new questions (\textbf{Bottom}).
    }
    \label{fig:figure_architecture}
\end{figure*}

\subsection{GNN Architecture}
\label{main:architecture:details}

Given the graph representation, we define \ours as a link prediction model trained end-to-end.
As illustrated in \Cref{fig:figure_architecture}, an encoder performs relation-aware message passing to produce node embeddings for subgroups, individuals, and choices,
and the decoder performs link prediction from output node embeddings.
To generalize to \textit{new questions} (setting 3) whose choice nodes have no edges at test time,
we additionally train an LLM-to-GNN projection that maps choice nodes' text features (frozen LLM hidden states) to representations in the GNN output embedding space.

\vspace{-8pt}
\paragraph{Input node features.}
Individual nodes $\mathcal{U}$ are non-identifiable, thereby assigned a uniform feature $\smash{Z_{\mathcal{U}} = \mathbf{1}_{|\mathcal{U}|}}$.
For subgroup nodes $\mathcal{S}$, we learn input node features via a learnable table $\smash{Z_{\mathcal{S}} \in \mathbb{R}^{|\mathcal{S}|\times d_{\mathcal{S}}}}$, with a feature dimension $d_{\mathcal{S}}$.
For choice nodes $\mathcal{C}$,
we also maintain a learnable table $\smash{Z_{\mathcal{C}} \in \mathbb{R}^{|\mathcal{C}|\times d_{\mathcal{C}}}}$ with a feature dimension $d_{\mathcal{C}}$;
no textual information about choices are provided as input to the graph.

\vspace{-8pt}
\paragraph{Graph encoder.}
We adopt heterogeneous graph extensions of GNNs, e.g., RGCN, GAT, and GraphSAGE \citep{schlichtkrull2018modeling, velivckovic2017graph, hamilton2017inductive}.
Let $\smash{z_w^{(0)}}$ be the input feature for node $w$ from $Z_{\mathcal{U}}$, $Z_{\mathcal{S}}$, or $Z_{\mathcal{C}}$.
An $L$-layer graph encoder computes
\begin{equation}
\begin{aligned}
z_w^{(\ell+1)}
&= \sigma\Bigg(
\sum_{r\in\mathcal{R}}
\Big[
\operatorname*{AGG_{\mathit{r}}}_{v \in \mathcal{N}_r(w)}
\;\phi_{r}^{(\ell)}\!\big(z_w^{(\ell)},\, z_v^{(\ell)}\big)
\Big] \\
&\qquad\;\; + \phi_{\text{self}}^{(\ell)}\big(z_w^{(\ell)}\big)
\Bigg), \ell = 0,\dots,L-1 .
\end{aligned}
\label{eq:encoder}
\end{equation}
where $\smash{\mathcal{R}=\{\mathcal{U\rightarrow S},\mathcal{S\rightarrow U},\mathcal{U\rightarrow C},\mathcal{C\rightarrow U}\}}$
are types of two bidirectional relations (membership and response),
$\smash{\phi_{r}^{(\ell)}}$ is a relation-specific message passing,
$\smash{\phi_{\text{self}}^{(\ell)}}$ is a self-loop,
$\operatorname*{AGG_{\mathit{r}}}$ is a per-relation aggregation over neighbors $\mathcal{N}_r(w)$,
and $\sigma$ is a non-linear activation function.
We present the details of each function for different GNN architectures in \Cref{appendix:training_details}.
After the final layer $L$,
we apply a node-type-specific linear projection on $\smash{z_w^{(L)}}$
to obtain the output embedding $z_w^{O} \in \mathbb{R}^{d_{\text{GNN}}}$, where $d_{\text{GNN}}$ is the dimension of output node embeddings.

\vspace{-8pt}
\paragraph{Link prediction decoder.}
The decoder consists of a dot-product and a softmax classifier.
For an individual $u \in \mathcal{U}$ and a question $q$ with a set of choice nodes $\mathcal{C}_q\subseteq \mathcal{C}$,
the score between the individual and each choice $c \in \mathcal{C}_q$ is obtained as $\operatorname{Dot}(u,c) = (z^O_u)^\top z^O_c$.
These scores are then converted to a distribution over choices, with a learnable temperature $\tau$:
\begin{equation}
p(c\,|\,u,q) \;=\; 
\frac{\exp\big(\operatorname{Dot}(u,c)/\tau\big)}{\sum_{c'\in \mathcal{C}_q} \exp\big(\operatorname{Dot}(u,c')/\tau\big)}.
\label{eq:softmax}
\end{equation}

\vspace{-8pt}
\paragraph{LLM-to-GNN representation map.}
In setting 3 (new questions),
choice nodes for the new question are isolated in the graph since we do not have any responses for that question yet,
and have no learned features in the learnable table $Z_{\mathcal{C}}$.
Therefore, the graph encoder cannot produce the output embedding for new choice nodes.
To enable prediction, we generate a substitute embedding directly from its text features by learning an LLM-to-GNN representation mapping on seen questions.
For a choice $c$,
the mapping takes a language representation of the choice's text features (a frozen LLM's hidden state $h_{\text{LLM}}(c) \in \mathbb{R}^{d_{\text{LLM}}}$)
then outputs $z'_c=\mathbf{W}_{\text{proj}}\,h_{\text{LLM}}(c) \in \mathbb{R}^{d_{\text{GNN}}}$,
where $d_{\text{LLM}}$ and $d_{\text{GNN}}$ are dimensions of LLM hidden states and GNN output embeddings, respectively.

The projection is trained on seen choice nodes by matching $z'_c$ to the output node embedding $z_c^O$, inspired by previous work \citep{sheng2025language, zhang2019attributed}.
At inference time, for a new question $q$, we compute $z'_c$ for each $c\in\mathcal{C}_q$ and plug these into the decoder in place of $z_c^O$.
We note that this mapping is only needed in setting~3; settings~1–2 use the output embeddings $z_c^O$ directly.
Furthermore, since we use frozen LLM hidden states, we only add $d_{\text{LLM}} \times d_{\text{GNN}}$ trainable parameters (far less than LLM fine-tuning) and do not require querying an LLM each time to predict new edges, unlike prompt-based LLM approaches.

\subsection{Training objective}
\label{main:architecture:learning_objectives}

\paragraph{Link prediction.}
Following self-supervised link prediction \citep{kipf2016variational, berg2017graph},
we train end-to-end by exposing a subset of train edges to the graph encoder and supervising the model to reconstruct the rest.
At each train step we randomly mask response edges from $E_{\mathcal{UC}}$,
with a masking strategy defined in Section~\ref{main:experiment} per setting.
For example, say we masked a response edge $(u,c)$ for an individual $u$ and a choice $c$ where $c$ belongs to a question $q(c)$.
The decoder generates a probability $p(c\,|\,u,q(c))$ by \Cref{eq:softmax}.
We aim to minimize the cross-entropy loss
\begin{equation}
-\!\!\sum_{(u,c)\in \text{masked}}\!
\log p(c\,|\,u,q(c))
\label{eq:linkpred}
\end{equation}
It requires no explicit negative sampling:
the masked response edge $(u,c)$ is the positive edge, while $(u,c')$ for all $c' \in \mathcal{C}_{q(c)} \setminus \{c\}$ act as implicit negatives through the normalization with softmax.

\vspace{-8pt}
\paragraph{LLM-to-GNN map.}
For setting 3 (new question), we learn a linear mapping $\mathbf{W}_{\text{proj}}$ by minimizing
\begin{equation}
\sum_{c \in \mathcal{C}_{\text{train}}}\bigl\|\mathbf{W}_{\text{proj}}h_{\text{LLM}}(c)-z^O_c\bigr\|_2^2
+ \alpha\|\mathbf{W}_{\text{proj}}\|_2^2
\label{eq:lproj}
\end{equation}
where $\mathcal{C}_{\text{train}}$ is the set of choice nodes available during training,
$h_{\text{LLM}}$ is a frozen LLM's hidden state for a text feature of a choice node $c$,
and $z_c^O$ is the output embedding of an $L$-layer graph encoder.
$\alpha$ is a hyperparameter of a ridge regression selected by the prediction accuracy on the validation set.

%% file: main_sections/experiments.tex
\section{Experiments}
\label{main:experiment}

\subsection{Experimental Setup}
\label{main:experiment:setup}

\textbf{Datasets.}
We evaluate on three simulation datasets:
(1) \textsc{OpinionQA} public opinion polls \citep{santurkar2023whose},
comprising responses from 76K individuals to 500 questions spanning various social topics (e.g., political attitudes, media consumption);
(2) \textsc{Twin‑2K} \citep{toubia2025twin},
a 150‑item battery including economic preferences, cognitive biases, and personality traits, administered to 2K individuals; and
(3) \textsc{Dunning-Kruger} effect replication \citep{jansen2021rational, binz2025foundation},
20 grammar and logical reasoning questions with pre- and post-question confidence ratings, administered on 3K individuals.
For (1) and (2), nine demographic attributes (e.g., age, gender) are used to define individual features and subgroup nodes;
for (3), responses to four pre-question confidence ratings are used as individual features.
Examples of questions and choices are in \Cref{appendix:question_examples}.
Dataset splits are described per setting below;
graph statistics appear in \Cref{appendix:graph_statistics}.

\vspace{-8pt}
\paragraph{Evaluation metric.}
We use average prediction accuracy, comparing the individual's true choice to the highest-probability model prediction.
Specifically, for a test response edge $(u,c)$ with a question $q(c)$ that $c$ belongs to,
the model's prediction is correct if $c = \smash{\operatorname{argmax}_{c' \in \mathcal{C}_{q(c)}}p(c' | u, q(c))}$ (\Cref{eq:softmax}) and incorrect otherwise.

\vspace{-8pt}
\paragraph{Compared methods.}
We compare \ours against five LLM-based baselines and lower/upper performance bounds.
Prompt examples for each of the baselines are provided in \Cref{appendix:prompts_llm}.
\vspace{-5pt}
\begin{description}[leftmargin=3pt,labelindent=3pt,itemsep=2pt,parsep=0pt]
    \item[1.] Zero‑shot prompting: Prompt with individual features, following \citet{santurkar2023whose}.
    \item[2.] Few‑shot prompting: Prompt with individual features and the individual's prior responses, following \citet{hwang2023aligning, kim2025few}.
    \item[3.] Agentic CoT prompting: A chain-of-thought (CoT) framework consisting of a reflection agent and a prediction agent \citep{park2024generative}.
    \item[4.] Supervised fine‑tuning (SFT): Fine‑tune an LLM to predict the answer token given individual features \citep{cao2025specializing, suh2025language, yaono, kolluri2025finetuningllmshumanbehavior}.
    \item[5.] Few‑shot fine‑tuning (Few‑shot FT): Fine‑tune an LLM with individual features and the individual's prior responses \citep{zhaogroup}.
    \item[6.] Random (lower bound): Uniformly sample a choice from the question’s available options.
    \item[7.] Human retest (upper bound): When available from dataset authors, report test-retest accuracy.
    It is the probability that the same individual repeats the same choice when re‑asked the same question after a fixed time interval (e.g., two weeks).
\end{description}
\vspace{-5pt}
For main experiments we adopt three instruction-tuned language models, LLaMA-2-7B, Mistral-7B-v0.1, Qwen3-8B \citep{touvron2023llama, jiang2023mistral, yang2025qwen3}.
We also present additional inference results in \Cref{appendix:additional_experiment}
including state-of-the-art proprietary models.
We note that results with additional models do not alter the overall trends or our main conclusions.

\subsection{Setting 1: Missing Responses (Imputation)}
\label{main:experiment:setting1}

\input{table/table_setting_1}

\textbf{Setup.}
We follow the split scheme of \cite{zhaogroup}:
each dataset is first split at an individual level into 35/5/60\% train/validation/test individuals.
For each individual held out for validation/test, 40\% of their responses 
are also available during training, while 60\% are held out for evaluation.
LLM fine-tuning prompts and train graphs are built upon all responses from the train individuals and 40\% responses from the validation/test individuals.
At each training step of \ours,
50\% of response edges in the train graph are randomly masked and used as supervision edges, while all membership edges and unmasked train response edges serve as message passing edges.
At validation/test, the entire train graph is used for message passing to predict held-out edges.
For LLM few-shot prompts we use 3 or 8 in-context examples,
selected from training data by the highest cosine similarity of text embeddings, following \citet{liu2021makes} and \citet{hwang2023aligning}.
Please refer to \Cref{appendix:training_details} for additional details.

\vspace{-8pt}
\paragraph{Results.}
\Cref{table:table_transductive_individual} reports results.
\ours outperforms all LLM prompting baselines and SFT, and matches the strongest LLM baseline, 8-shot fine-tuning.
Performance of LLM-based methods generally improves with more sophisticated prompt design and compute, from zero-shot prompting to few-shot fine-tuning;
however, \ours attains comparable accuracy without using any textual features, relying solely on a learnable feature table of choices and subgroups.
We attribute this to the relational structure that alone provides sufficient signal about the choice, even in the absence of textual information.
For example, GNN infers that two frequently co-selected choices share latent attributes and therefore someone choosing one will likely choose the other, even without text of choices.
Taken together, these results highlight the value of relational structure for accurate prediction.

\subsection{Setting 2: New Individuals}
\label{main:experiment:setting2}

\input{table/table_setting_3}

\textbf{Setup.}
The split is also done at an individual level: \(35\%\) train, \(5\%\) validation, and \(60\%\) test individuals.
In contrast to setting 1 where we hold out 60\% responses from each validation/test individual, here we hold out all responses to have a new individual at test time.
We also modify \ours training to teach the model how to make predictions for new individuals.
At each training step, we randomly select 50\% of training individuals, mask all of their response edges to use as supervision edges,
and use all membership edges and unselected training individuals' response edges for message passing.

\vspace{-8pt}
\paragraph{Results.}
\Cref{table:table_inductive_individual} reports results for setting 2.
\ours outperforms the LLM prompting baselines and matches the strongest LLM baseline, SFT.
Trends mirror that of setting 1:
(i) zero‑shot prompting and CoT exceed a random baseline and benefit from stronger LLMs but fall behind SFT, and
(ii) fine‑tuning narrows performance gaps across LLM families.
In \ours, a new individual connects only to subgroup nodes via membership edges;
with input features $\mathbf{1}_{|\mathcal{U}|}$ (\Cref{main:architecture:details}),
their output node embeddings are obtained entirely by aggregating messages from subgroup neighbors (\Cref{eq:encoder}).
By masking out all response edges for 50\% individuals during training,
the learnable subgroup features \(Z_{\mathcal{S}}\) are encouraged to encode representations that generalize to new individuals, precisely what is needed for the current setting.
LLM‑based methods can acquire similar knowledge by iteratively seeing pairs of individual features and responses, but at a substantially higher computational cost.

\subsection{Setting 3: New Questions}
\label{main:experiment:setting3}

\textbf{Setup.}
We split at the question level into 70/10/20\% train/validation/test, following \citet{suh2025language}.
Validation/test questions are entirely unseen during training.
Even at test time their choice nodes are isolated in the graph, and do not have learned input features in the table \(Z_{\mathcal{C}}\).
Responses to train questions from all individuals are used to fine‑tune LLMs or to construct the train graph.
At validation/test, responses to train questions are used as few-shot examples or as message‑passing edges.

\ours is trained in two stages.
In the first stage, we train the GNN using the link prediction objective (\Cref{eq:linkpred}) to learn representations of individual and choice nodes in the train graph.
To this end, we initially hold out a small fraction (5\%) of response edges from the train graph, which we call ``transductive validation edges''.
At each GNN training step, remaining 95\% response edges in the graph are partitioned into 50\% supervision edges and 50\% message-passing edges as done in setting 1.
At the GNN checkpoint with the best accuracy on the transductive validation edges, we extract the output node embeddings $z_c^O$ of choice nodes $c \in \mathcal{C}_{\text{train}}$ (\Cref{eq:encoder}).
In the second stage, we train a linear projection to map LLM hidden states of $\mathcal{C}_{\text{train}}$ text features to the GNN output node embedding space (\Cref{eq:lproj}).
At test time, we make predictions for new questions using the projected embeddings of their choices, as described in Section~\ref{main:architecture:details}.

\vspace{-8pt}
\paragraph{Results.}
\Cref{table:table_inductive_question} reports results for setting 3.
\ours, with the LLM-to-GNN representation mapping, remains competitive with all LLM-based methods, while requiring far fewer trainable parameters.
This performance is achieved with LLM hidden states -- GNN output embedding pairs from 500 (\textsc{OpinionQA}), 150 (\textsc{Twin-2K}), or 20 (\textsc{Dunning-Kruger}) questions.
We also observe that the choice of LLM affects \ours:
accuracies achieved with \ours correlate with those of the corresponding LLM‑based baselines,
indicating that gains from stronger LLMs translate through the mapping \citep{sheng2025language}.

\subsection{Error Analysis}
\label{sec:complementarity}

To better understand the relationship between GEMS and LLM predictions,
we compute contingency tables between the best-performing LLM method and the best-performing GEMS variant on the \textsc{OpinionQA} dataset across all three settings.

\begin{table}[H]
    \centering
    \small
    \setlength{\tabcolsep}{2.5pt}
    \caption{
        Contingency tables (\%) between the best LLM method and GEMS on \textsc{OpinionQA}. \checkmark (\ding{55}) indicates that the model makes a correct (incorrect) prediction.
    }    
    \begin{tabular}{ll cc cc cc}
    \toprule
    & & \multicolumn{2}{c}{\textbf{Setting 1}} & \multicolumn{2}{c}{\textbf{Setting 2}} & \multicolumn{2}{c}{\textbf{Setting 3}} \\
    \cmidrule(lr){3-4} \cmidrule(lr){5-6} \cmidrule(lr){7-8}
    & & LLM\checkmark & LLM\ding{55} & LLM\checkmark & LLM\ding{55} & LLM\checkmark & LLM\ding{55} \\
    \midrule
    GNN\checkmark && 53.4 & 3.6 & 47.9 & 2.8 & 40.9 & 9.3 \\
    GNN\ding{55}  && 3.4 & 39.6 & 2.6 & 46.7 & 11.0 & 38.8 \\
    \bottomrule
    \end{tabular}
    \label{tab:contingency}
\end{table}
\vspace{-5pt}

Across all three settings, shared failures (bottom right) are substantially larger than complementary subsets (off-diagonal entries),
indicating that two methods tend to make similar predictions.
In Settings~1 and~2, the complementary subsets where only one model succeeds are small (3--4\% each), suggesting limited headroom for simple cross-model ensembling.
In Setting~3, the off-diagonal mass is larger (${\sim}$20\% combined), implying greater complementarity between \ours and LLM-based methods when generalizing to new questions.

\subsection{Comparison with Classical Baselines}
\label{sec:classical}

The previous sections show that \ours matches or outperforms LLM-based methods.
To isolate the contribution of graph structure beyond classical tabular or factorization methods, we compare \ours against two additional baselines.

\noindent \textbf{XGBoost} (Settings 1 and 2).
We train a per-question XGBoost classifier whose input is a one-hot encoding of individual features (e.g., demographic traits)
and whose output is a distribution over the available choices.
XGBoost cannot generalize to entirely new questions (Setting~3), as it lacks a mechanism to transfer across questions.
 
\noindent \textbf{Matrix factorization} (Setting 1).
We learn embeddings for each individual and each choice by optimizing a cross-entropy loss over softmax-normalized dot products.
This method applies only to Setting~1, as it cannot construct embeddings for new individuals or new questions at test time.

Results are reported in Table~\ref{tab:classical}.
In Setting~1, where the model must capture relational structure among discrete choices across multiple questions, matrix factorization underperforms both GEMS and LLM few-shot fine-tuning, with XGBoost performing even worse.
In Setting~2, which relies more heavily on individual features, the simpler baselines perform more competitively, though GEMS still achieves a consistent edge.
Importantly, neither XGBoost nor matrix factorization can handle Setting~3, while GEMS can.
These results suggest that GEMS' primary advantage over classical methods lies in its ability to learn richer relational structure through message passing.

\begin{table}[H]
\centering
\small
\caption{
    Comparison with classical methods on \textsc{OpinionQA} (OQ), \textsc{Twin-2K} (TW), and \textsc{Dunning-Kruger} (DK).
    We note that neither XGBoost nor matrix factorization can handle Setting~3.
}
\begin{tabular}{lccc}
\toprule
\textbf{Method} & \textbf{OQ} & \textbf{TW} & \textbf{DK} \\
\midrule
\multicolumn{4}{l}{\textit{Setting 1: Missing Responses}} \\
\midrule
XGBoost & 50.20 & 61.94 & 56.44 \\
Matrix Factorization & 52.12 & 62.77 & 56.53 \\
LLM Best (Few-shot FT, 8) & 56.76 & 66.36 & 57.21 \\
GEMS Best & \textbf{57.00} & \textbf{66.62} & \textbf{57.89} \\
\midrule
\multicolumn{4}{l}{\textit{Setting 2: New Individuals}} \\
\midrule
XGBoost & 50.16 & 61.61 & 56.24 \\
LLM Best (SFT) & 50.49 & 61.85 & 56.66 \\
GEMS Best & \textbf{50.73} & \textbf{62.50} & \textbf{57.07} \\
\bottomrule
\end{tabular}
\label{tab:classical}
\end{table}

%% file: table/table_setting_1.tex
\begin{table*}[!ht]
    \centering
    \scriptsize
    \caption{
Accuracy of imputing missing responses.
Numbers indicate mean test accuracy with standard deviation from 3 train/val/test random splits with different seeds.
Numbers in parentheses indicate the number of in-context examples (e.g., 3) or the GNN architecture (e.g., RGCN).
For each dataset, bold marks the best accuracy per \ours and LLM-based methods; underline for the runner-up.
N.A. and C.L. stand for `not available' and `context limit', respectively.
}
    \label{table:table_transductive_individual}
    \resizebox{\textwidth}{!}{
    {\setlength{\tabcolsep}{2.5pt}
    \begin{tabular}{c|ccc|ccc|ccc}
    \toprule
    & \multicolumn{3}{c|}{\textsc{OpinionQA}}
    & \multicolumn{3}{c|}{\textsc{Twin-2K}}
    & \multicolumn{3}{c}{\textsc{Dunning-Kruger}} \\
    \multirow{1}{*}{\textbf{Methods}}
    & \textbf{LLaMA2-7B} & \textbf{Mistral-7B} & \textbf{Qwen3-8B}
    & \textbf{LLaMA2-7B} & \textbf{Mistral-7B} & \textbf{Qwen3-8B}
    & \textbf{LLaMA2-7B} & \textbf{Mistral-7B} & \textbf{Qwen3-8B} \\
    \midrule
    Random & \multicolumn{3}{c|}{27.87}             & \multicolumn{3}{c|}{35.05} & \multicolumn{3}{c}{20.00} \\
    Human retest & \multicolumn{3}{c|}{\text{N.A.}} & \multicolumn{3}{c|}{81.72} & \multicolumn{3}{c}{N.A.}  \\
    \midrule
    Zero-shot
    & 29.18\verytiny{$\pm$0.15} & 34.63\verytiny{$\pm$0.19} & 39.38\verytiny{$\pm$0.20}
    & 41.49\verytiny{$\pm$0.31} & 42.47\verytiny{$\pm$0.27} & 52.06\verytiny{$\pm$0.38}
    & 22.76\verytiny{$\pm$0.59} & 39.00\verytiny{$\pm$0.73} & 41.82\verytiny{$\pm$0.41} \\    
    \midrule
    Few-shot (3)
    & 38.54\verytiny{$\pm$0.21} & 42.52\verytiny{$\pm$0.06} & 42.21\verytiny{$\pm$0.08}
    & 41.44\verytiny{$\pm$0.88} & 48.25\verytiny{$\pm$0.73} & 54.10\verytiny{$\pm$0.51}
    & 26.77\verytiny{$\pm$0.23} & 43.54\verytiny{$\pm$0.72} & 46.81\verytiny{$\pm$0.48} \\
    Few-shot (8)
    & 37.91\verytiny{$\pm$0.65} & 45.78\verytiny{$\pm$0.56} & 43.66\verytiny{$\pm$0.59}
    & 43.40\verytiny{$\pm$0.99} & 51.26\verytiny{$\pm$0.84} & 56.08\verytiny{$\pm$1.01}
    & 26.34\verytiny{$\pm$0.51} & 43.59\verytiny{$\pm$0.60} & 47.60\verytiny{$\pm$0.67} \\
    \midrule
    Agentic CoT (3)
    & 32.19\verytiny{$\pm$0.25} & 41.37\verytiny{$\pm$0.47} & 47.63\verytiny{$\pm$0.17}
    & 33.13\verytiny{$\pm$1.57} & 50.14\verytiny{$\pm$0.93} & 57.89\verytiny{$\pm$1.80}
    & 31.01\verytiny{$\pm$1.51} & 34.41\verytiny{$\pm$2.84} & 51.18\verytiny{$\pm$0.79} \\
    Agentic CoT (8)
    & 28.80\verytiny{$\pm$0.15} & 38.43\verytiny{$\pm$0.31} & 47.97\verytiny{$\pm$0.36} 
    & \text{C.L.}               & 48.76\verytiny{$\pm$0.53} & 60.20\verytiny{$\pm$1.28}
    & 31.68\verytiny{$\pm$0.88} & 35.45\verytiny{$\pm$1.23} & 54.71\verytiny{$\pm$1.27} \\
    \midrule
    SFT
    & 49.41\verytiny{$\pm$0.12} & 50.56\verytiny{$\pm$0.14} & 48.84\verytiny{$\pm$0.14}
    & 61.23\verytiny{$\pm$0.13} & 61.85\verytiny{$\pm$0.13} & 61.49\verytiny{$\pm$0.15}
    & 56.47\verytiny{$\pm$0.10} & 56.45\verytiny{$\pm$0.14} & 56.47\verytiny{$\pm$0.03} \\
    \midrule
    Few-shot FT (3)
    & 55.59\verytiny{$\pm$0.11} & \underline{56.31\verytiny{$\pm$0.10}} & 55.09\verytiny{$\pm$0.14}
    & 63.51\verytiny{$\pm$0.15} & 63.91\verytiny{$\pm$0.16}             & 62.61\verytiny{$\pm$0.19}
    & 56.81\verytiny{$\pm$0.15} & 56.88\verytiny{$\pm$0.05}             & 56.71\verytiny{$\pm$0.12} \\
    Few-shot FT (8)
    & 55.98\verytiny{$\pm$0.12}             & \textbf{56.76}\verytiny{$\pm$\textbf{0.13}} & 55.61\verytiny{$\pm$0.13}
    & \underline{65.86\verytiny{$\pm$0.17}} & \textbf{66.36}\verytiny{$\pm$\textbf{0.13}} & 65.27\verytiny{$\pm$0.16}
    & \underline{57.18\verytiny{$\pm$0.02}} & \textbf{57.21}\verytiny{$\pm$\textbf{0.41}} & 56.94\verytiny{$\pm$0.11} \\
    \midrule
    \midrule
    \ours (RGCN)
    & \multicolumn{3}{c|}{\underline{56.89\verytiny{$\pm$0.12}}}
    & \multicolumn{3}{c|}{\underline{66.36\verytiny{$\pm$0.13}}}
     & \multicolumn{3}{c}{\underline{57.68\verytiny{$\pm$0.13}}} \\
    \ours (GAT)
    & \multicolumn{3}{c|}{56.40\verytiny{$\pm$0.10}}
    & \multicolumn{3}{c|}{66.01\verytiny{$\pm$0.14}}
     & \multicolumn{3}{c}{56.95\verytiny{$\pm$0.09}} \\
        \ours (SAGE)
    & \multicolumn{3}{c|}{\textbf{57.00}\verytiny{$\pm$\textbf{0.12}}}
    & \multicolumn{3}{c|}{\textbf{66.62}\verytiny{$\pm$\textbf{0.12}}}
     & \multicolumn{3}{c}{\textbf{57.89}\verytiny{$\pm$\textbf{0.10}}} \\
    \bottomrule
    \end{tabular}
    }
    }
\end{table*}

\begin{table*}[!t]
    \centering
    \scriptsize
    \caption{
    Accuracy of predicting responses of new, unseen individuals.
    Numbers indicate mean test accuracy with standard deviation from 3 train/val/test random splits with different seeds.
    For LLM, few-shot methods are not applicable since prior responses for new individuals are not available.
    }
    \label{table:table_inductive_individual}
    \resizebox{\textwidth}{!}{
    {\setlength{\tabcolsep}{2.5pt}
    \begin{tabular}{c|ccc|ccc|ccc}
    \toprule
    & \multicolumn{3}{c|}{\textsc{OpinionQA}}
    & \multicolumn{3}{c|}{\textsc{Twin-2K}}
    & \multicolumn{3}{c}{\textsc{Dunning-Kruger}} \\
    \multirow{1}{*}{\textbf{Methods}}
    & \textbf{LLaMA2-7B} & \textbf{Mistral-7B} & \textbf{Qwen3-8B}
    & \textbf{LLaMA2-7B} & \textbf{Mistral-7B} & \textbf{Qwen3-8B}
    & \textbf{LLaMA2-7B} & \textbf{Mistral-7B} & \textbf{Qwen3-8B} \\
    \midrule
    Random & \multicolumn{3}{c|}{27.87} & \multicolumn{3}{c|}{35.05} & \multicolumn{3}{c}{20.00} \\
    \midrule
    Zero-shot
    & 29.15\verytiny{$\pm$0.15} & 34.40\verytiny{$\pm$0.13} & 38.97\verytiny{$\pm$0.16}
    & 41.57\verytiny{$\pm$0.39} & 43.03\verytiny{$\pm$0.50} & 51.79\verytiny{$\pm$0.27}
    & 22.44\verytiny{$\pm$0.39} & 38.83\verytiny{$\pm$0.70} & 42.06\verytiny{$\pm$0.41} \\
    \midrule
    Agentic CoT
    & 18.44\verytiny{$\pm$0.47} & 33.84\verytiny{$\pm$0.31} & 39.53\verytiny{$\pm$0.22}
    & 21.91\verytiny{$\pm$0.82} & 45.30\verytiny{$\pm$0.34} & 53.45\verytiny{$\pm$0.43}
    & 31.52\verytiny{$\pm$1.47} & 34.04\verytiny{$\pm$0.66} & 49.96\verytiny{$\pm$1.19} \\
    \midrule
    SFT
    & \underline{49.35\verytiny{$\pm$0.15}} & \textbf{50.49}\verytiny{$\pm$\textbf{0.17}} & 48.87\verytiny{$\pm$0.16}
    & 61.29\verytiny{$\pm$0.22} & \textbf{61.85}\verytiny{$\pm$\textbf{0.19}} & \underline{61.38\verytiny{$\pm$0.22}}
    & \underline{56.54\verytiny{$\pm$0.33}} & \textbf{56.66}\verytiny{$\pm$\textbf{0.17}} & 56.50\verytiny{$\pm$0.34} \\
    \midrule
    \midrule
    \ours (RGCN)
    & \multicolumn{3}{c|}{\underline{50.50\verytiny{$\pm$0.12}}}
    & \multicolumn{3}{c|}{\underline{62.39\verytiny{$\pm$0.14}}}
     & \multicolumn{3}{c}{\underline{56.76\verytiny{$\pm$0.21}}} \\
    \ours (GAT)
    & \multicolumn{3}{c|}{50.36\verytiny{$\pm$0.14}}
    & \multicolumn{3}{c|}{62.22\verytiny{$\pm$0.14}}
     & \multicolumn{3}{c}{56.70\verytiny{$\pm$0.10}} \\
    \ours (SAGE)
    & \multicolumn{3}{c|}{\textbf{50.73}\verytiny{$\pm$\textbf{0.11}}}
    & \multicolumn{3}{c|}{\textbf{62.50}\verytiny{$\pm$\textbf{0.19}}}
     & \multicolumn{3}{c}{\textbf{57.07}\verytiny{$\pm$\textbf{0.32}}} \\
    \bottomrule
    \end{tabular}
    }
    }
\end{table*}

%% file: table/table_setting_3.tex
\begin{table*}[!t]
    \centering
    \scriptsize
    \caption{
    Accuracy of predicting human responses to new, unseen questions.
    Numbers indicate mean test accuracy with standard deviation from 3 train/val/test random splits with different seeds.
    For \ours, within a row, each column indicates the performance with hidden states from different LLMs.
    Experimental details are in \Cref{appendix:training_details}.
    }
    \label{table:table_inductive_question}
    \resizebox{\textwidth}{!}{
    {\setlength{\tabcolsep}{2.5pt}
    \begin{tabular}{c|ccc|ccc|ccc}
    \toprule
    & \multicolumn{3}{c|}{\textsc{OpinionQA}}
    & \multicolumn{3}{c|}{\textsc{Twin-2K}}
    & \multicolumn{3}{c}{\textsc{Dunning-Kruger}} \\
    \multirow{1}{*}{\textbf{Methods}}
    & \textbf{LLaMA2-7B} & \textbf{Mistral-7B} & \textbf{Qwen3-8B}
    & \textbf{LLaMA2-7B} & \textbf{Mistral-7B} & \textbf{Qwen3-8B}
    & \textbf{LLaMA2-7B} & \textbf{Mistral-7B} & \textbf{Qwen3-8B} \\
    \midrule
    Random & \multicolumn{3}{c|}{27.87} & \multicolumn{3}{c|}{35.05} & \multicolumn{3}{c}{20.00} \\
    \midrule
    Zero-shot
    & 29.15\verytiny{$\pm$0.57} & 35.60\verytiny{$\pm$2.91} & 38.84\verytiny{$\pm$1.08}
    & 40.03\verytiny{$\pm$2.45} & 41.30\verytiny{$\pm$3.69} & 50.94\verytiny{$\pm$2.76}
    & 19.77\verytiny{$\pm$5.64} & 43.87\verytiny{$\pm$4.74} & 44.12\verytiny{$\pm$4.40} \\
    \midrule
    Few-shot (3)
    & 37.93\verytiny{$\pm$2.24} & 42.49\verytiny{$\pm$3.16} & 42.74\verytiny{$\pm$2.87}
    & 42.09\verytiny{$\pm$3.38} & 47.88\verytiny{$\pm$1.93} & 54.02\verytiny{$\pm$4.19}
    & 23.83\verytiny{$\pm$5.85} & 49.31\verytiny{$\pm$4.48} & 52.55\verytiny{$\pm$3.71} \\
    Few-shot (8)
    & 37.98\verytiny{$\pm$1.62} & 42.81\verytiny{$\pm$3.39} & 44.05\verytiny{$\pm$2.65}
    & 41.15\verytiny{$\pm$2.77} & 47.93\verytiny{$\pm$2.30} & 55.09\verytiny{$\pm$2.50}
    & 23.81\verytiny{$\pm$6.30} & 44.30\verytiny{$\pm$4.12} & 52.93\verytiny{$\pm$3.97} \\
    \midrule
    Agentic CoT (3)
    & 31.46\verytiny{$\pm$2.92} & 40.20\verytiny{$\pm$1.60} & 45.90\verytiny{$\pm$3.57}
    & 32.16\verytiny{$\pm$3.66} & 49.67\verytiny{$\pm$3.61} & 56.18\verytiny{$\pm$2.74}
    & 23.56\verytiny{$\pm$18.8} & 31.18\verytiny{$\pm$8.58} & 53.54\verytiny{$\pm$4.41} \\ 
    Agentic CoT (8)
    & 27.15\verytiny{$\pm$1.42} & 37.45\verytiny{$\pm$4.94} & 46.18\verytiny{$\pm$3.70}
    & \text{C.L.}               & 48.24\verytiny{$\pm$5.43} & 58.08\verytiny{$\pm$2.83}
    & 24.45\verytiny{$\pm$17.8} & 30.98\verytiny{$\pm$9.37} & \underline{53.72\verytiny{$\pm$4.61}} \\
    \midrule
    SFT
    & 44.12\verytiny{$\pm$2.30} & 47.86\verytiny{$\pm$0.95} & 43.95\verytiny{$\pm$0.87}
    & 55.85\verytiny{$\pm$1.21} & 56.21\verytiny{$\pm$0.96} & 56.24\verytiny{$\pm$1.42}
    & 33.65\verytiny{$\pm$13.1} & 48.12\verytiny{$\pm$7.46} & 43.08\verytiny{$\pm$5.31} \\
    \midrule
    Few-shot FT (3)
    & 49.83\verytiny{$\pm$1.53} & \underline{51.77\verytiny{$\pm$1.09}} & 49.59\verytiny{$\pm$0.84}
    & 58.07\verytiny{$\pm$1.86} & 59.86\verytiny{$\pm$1.52}             & 59.99\verytiny{$\pm$1.33}
    & 41.18\verytiny{$\pm$16.3} & 49.28\verytiny{$\pm$9.40}             & \textbf{54.13\verytiny{$\pm$4.98}} \\
    Few-shot FT (8)
    & 50.11\verytiny{$\pm$1.97} & \textbf{51.83}\verytiny{$\pm$\textbf{1.47}} & 50.00\verytiny{$\pm$1.00}
    & 59.87\verytiny{$\pm$1.35} & \textbf{60.84}\verytiny{$\pm$\textbf{1.40}} & \underline{60.48\verytiny{$\pm$1.79}}
    & 41.81\verytiny{$\pm$9.96} & 51.79\verytiny{$\pm$8.00}                   & 53.47\verytiny{$\pm$5.93} \\
    \midrule
    \midrule
    \ours (RGCN)
    & 48.94\verytiny{$\pm$1.71} & \textbf{50.13}\verytiny{$\pm$\textbf{1.85}} & 49.07\verytiny{$\pm$1.48}
    & 56.24\verytiny{$\pm$3.65} & \textbf{60.37}\verytiny{$\pm$\textbf{2.47}} & 59.59\verytiny{$\pm$4.42}
    & 47.20\verytiny{$\pm$1.66} & 47.31\verytiny{$\pm$9.13}                   & \textbf{52.52}\verytiny{$\pm$\textbf{3.91}} \\
    \ours (GAT)
    & 46.87\verytiny{$\pm$1.78} & 49.25\verytiny{$\pm$2.46} & 48.52\verytiny{$\pm$2.13}
    & 52.00\verytiny{$\pm$1.52} & 56.57\verytiny{$\pm$1.95} & 57.38\verytiny{$\pm$2.44}
    & 45.02\verytiny{$\pm$3.09} & 44.57\verytiny{$\pm$4.85} & 52.01\verytiny{$\pm$3.16} \\
    \ours (SAGE)
    & 47.29\verytiny{$\pm$1.89} & \underline{49.84\verytiny{$\pm$1.98}} & 49.09\verytiny{$\pm$1.80}
    & 54.06\verytiny{$\pm$4.47} & 58.56\verytiny{$\pm$2.43} & \underline{60.03\verytiny{$\pm$3.88}}
    & 46.17\verytiny{$\pm$2.59} & 47.43\verytiny{$\pm$8.60} & \underline{52.02\verytiny{$\pm$3.58}} \\
    \bottomrule
    \end{tabular}
    }
    }
\end{table*}

%% file: main_sections/advantages.tex
\section{Advantages of \ours}
\label{main:advantages}

\begin{figure}[H]
    \centering
    \includegraphics[width=0.95\linewidth]{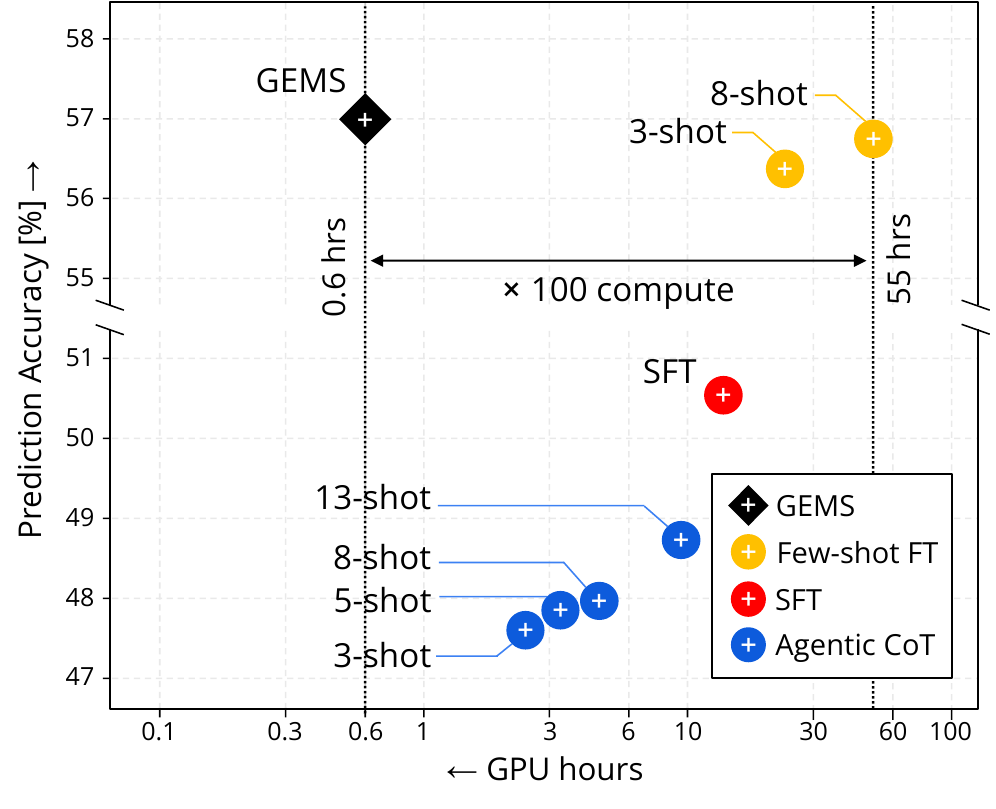}
    \caption{
    Accuracy \textit{vs.} GPU-hours (A100-80GB-SXM4) on the \textsc{OpinionQA} dataset, setting 1.
    Zero-/few-shot prompting accuracies fall below the plotted y-range.
    For LLM-based methods, we report the best result across three LLMs (LLaMA-2-7B, Mistral-7B-v0.1, and Qwen3-8B).
    For \ours, we report the best result across three models (RGCN, GAT, and SAGE).
    See \Cref{appendix:training_details} for details and \Cref{appendix:additional_experiment} for the extended figure.
    }
    \label{fig:figure_performance_compute}
\end{figure}

\noindent
While maintaining comparable accuracy to LLMs, \ours brings numerous practical advantages.

\vspace{-8pt}
\paragraph{Efficiency and scalability.}
As shown in \Cref{fig:figure_performance_compute},
\ours matches the strongest LLM-based methods while using $\sim\!10^2$ less compute and $\sim\!10^3$ fewer parameters
(see \Cref{appendix:training_details:model_size}), remaining tractable as both the base LLM and dataset scale.
Extrapolating from \Cref{fig:figure_performance_compute},
few-shot fine-tuning a 70B model on \textsc{OpinionQA} would require $\sim500$ GPU-hours,
whereas \ours completed in less than an hour.
Likewise, scaling to datasets $10\times$ larger than \textsc{OpinionQA} (e.g., SubPoP \citep{suh2025language})
would push an LLM fine-tuning toward $\sim\!10^3$ GPU-hours,
while \ours would train in a few hours.

This efficiency advantage also enables prediction ensembling \citep{lakshminarayanan2017simple},
which averages predicted probabilities across multiple models trained with different initializations.
Leveraging \ours' compute efficiency ($\sim\!10^2$ compared to LLM fine-tuning)
we ensemble predictions from 11 differently initialized GEMS models (still $\sim\!10$x cheaper than a single LLM fine-tuning) and observe consistent accuracy gains:

\begin{table}[H]
    \centering
    \small
    \caption{Ensembling results on \textsc{OpinionQA}. GEMS Ensembled averages predicted probabilities from 11 GEMS models trained with different initializations.}
    \resizebox{\linewidth}{!}{
    \begin{tabular}{lccc}
    \toprule
    \textbf{Method} & \textbf{Setting 1} & \textbf{Setting 2} & \textbf{Setting 3} \\
    \midrule
    LLM Best & 56.76 & 50.49 & 51.83 \\
    GEMS Best (single) & 57.00 & 50.73 & 50.13 \\
    GEMS Ensembled (11) & 57.31 & 50.86 & 51.65 \\
    \midrule
    Gain from ensembling & +0.31 & +0.13 & +1.52 \\
    \bottomrule
    \end{tabular}
    }
    \label{tab:ensemble}
\end{table}
\vspace{-8pt}

This demonstrates an advantage of \ours:
one can affordably train many models and ensemble predictions for improved accuracy, a strategy prohibitively expensive with LLM fine-tuning.

\vspace{-8pt}
\paragraph{Transparency and trustworthiness.}
LLMs are often trained on undisclosed data,
which creates contamination concerns where evaluation data (e.g., past behavioral studies) may have appeared during training \citep{deng2024unveiling}.
Furthermore, LLMs have been shown to display social biases in simulation, such as leaning towards certain groups' opinions \citep{santurkar2023whose}, stereotyping \citep{cheng2023compost}, or underestimating variance \citep{bisbee2024synthetic}.
Finally, pretrained LLMs are sensitive to prompt format \citep{lu2022fantastically,sclarquantifying},
with many formatting decisions involved in simulation tasks.
All of these issues
challenge the trustworthiness of LLM-based human simulations.

In contrast, \ours is trained from scratch on task‑specific data, removing issues of contamination or learning social biases from pretraining data.
Furthermore, there is no issue of ordering in-context examples or individual features, since
GNN aggregation is equivariant to the order of neighbors \citep{hamilton2017inductive}.
Prompt formatting is only relevant to \ours when the LLM-to-GNN mapping is used; even then, we find that it exhibits lower variance under prompt perturbations due to the training of the projection matrix.

\vspace{-8pt}
\paragraph{Insights.}
Finally, directly inspecting \ours' embeddings reveals insights about human behavior and preferences (Figure~\ref{fig:figure_embeddings}).
For example, when we train \ours on \textsc{OpinionQA}, we find that certain dimensions of opinions naturally emerge in the embedding space, with the first and second principal components corresponding to political ideology and class, respectively.
Second, even though we see clear subgroup-level patterns, inspecting the embeddings of individuals reveals substantial heterogeneity among individuals in the same subgroup.
These results emphasize the diversity of individuals beyond their demographics, in contrast with LLM methods that exhibit demographic stereotyping and underestimate variance within subgroups \citep{cheng2023compost,bisbee2024synthetic}.
Third, we find that \ours encodes nuanced meanings of questions that are missed by LLMs.
In particular, two choices that reflect similar ideology but have different wordings---for example, saying that ``reducing illegal immigration'' is ``a top priority'' and ``addressing climate change'' is ``not too important''---have similar \ours embeddings, while the LLM hidden states tend to be overly focused on surface word similarity (e.g., all ``a top priority'' are clustered regardless of the topic).
Please refer to \Cref{appendix:additional_experiment} for details.

%% file: main_sections/conclusion.tex
\section{Conclusion}
\label{main:conclusion}

We present \ours, a graph-based approach to model a large class of close-ended human simulation tasks previously dominated by LLMs.
By reformulating these tasks as link prediction on a graph, \ours learns from the relational structure of choices, complemented by a lightweight LLM‑to‑GNN projection.
We test three settings, including generalization to entirely new individuals or questions, and three datasets spanning opinion, cognitive, and educational domains.
Across these settings and datasets, \ours matches or surpasses the strongest LLM baselines, while offering nuanced insights and practical advantages.
Our work moves beyond the assumed dominance of LLMs for human simulation, offering lightweight alternatives without sacrificing predictive performance.

%% file: appendix_sections/limitations.tex
\newpage
\label{appendix:limitations_potential_risks}

\section{Limitations}
\label{appendix:limitations_potential_risks:limitations}

\vspace{-7pt}
\textbf{Limitations of the graph construction.}
In \Cref{fig:figure_1}, we encode individual features via subgroup nodes and connect individual nodes to subgroup nodes with membership edges.
The formulation is flexible:
it admits different subgroup granularities (e.g., intersectional groups),
alternative features (e.g., psychometrics test results),
or even peer–to-peer topology that links individuals by social ties as in social recommendation \citep{fan2019graph}.
However, in the datasets used here,
only basic demographic attributes \citep{santurkar2023whose, toubia2025twin}
and pre-question survey responses \citep{jansen2021rational} were available as individual features.
Exploring alternative graph constructions with richer features and analyzing their effects is an important future work.

\noindent
\textbf{Limitations of dataset coverage.}
Experiments use \textsc{OpinionQA}, \textsc{Twin-2K}, and a replication study of Dunning-Kruger effect, all participants from the United States.
Generalization to other countries or languages is untested.
We note, however, that \ours primarily learns from relational structure and uses language representations only when necessary (Setting~3),
making it less sensitive to the interface language than LLM-based simulation methods.
By contrast, prior works document that LLM performance can vary substantially by language;
this has been shown in public opinion simulation across countries \citep{qu2024performance} as well as in multilingual benchmarks \citep{singh2024global}.
Accordingly, \ours may offer robustness when linguistic variation is large, though this claim should be validated empirically with non‑English contexts.

\noindent
\textbf{Limitations of performance comparisons.}
Our LLM-based methods are fine‑tuned up to $\sim$10B parameters.
Larger models may further improve with fine‑tuning.
However, our experimental results show that after SFT or few‑shot fine-tuning, performance gaps across LLMs narrow (Table~\ref{table:table_transductive_individual}, \ref{table:table_inductive_individual}, \ref{table:table_inductive_question}),
indicating that \ours would remain competitive to fine-tuning larger LLMs.
Also, our compute figures (GPU hours, parameter counts) are not definitive in the sense that they vary with hardware, quantization, implementation of kernels, and model architecture details.
We still expect that the relative orders of compute figures would remain stable due to the significant difference in model sizes and input data formulation between LLMs and GNNs.
To support an informed comparison, we present the implementation details in \Cref{appendix:training_details}.

\noindent
\textbf{Insight claims.}
Dot‑product decoding based on output node embeddings makes the mechanics of prediction transparent (scores factor as similarities), but they are not causal explanations.
Qualitative inspection may risk being misread as normative judgments about groups.
Therefore, we suggest using them as diagnostic tools, complemented by ablations and sensitivity analyses.

\section{Ethical Considerations}
\label{appendix:limitations_potential_risks:ethical}

\vspace{-10pt}
\textbf{Privacy.}
The graph in \Cref{fig:figure_1} is constructed from de-identified individual features and response histories provided under the original data providers’ terms of use.
We neither collect nor store direct identifiers (e.g., names, addresses), and all analyses are performed on anonymized records.
To reduce identification risk, we report aggregate metrics (e.g., mean test accuracy) and do not release person-identifiable outputs.
For future work, we recommend treating individual-level data as sensitive, 
and adhering to applicable regulations, institutional review, and data-security best practices.

\noindent
\textbf{Responsible human simulation.}
Even when predictive accuracy is high, simulated responses must not replace human participants.
Human simulations should be carefully validated against historical data and deployed with guardrails, such as continued checks with human data, and with the goal of augmenting---not replacing---human studies.
Encoding people via demographic membership edges can inadvertently reinforce stereotypes or obscure within-group heterogeneity;
over-reliance on subgroup signals risks reproducing historical biases rather than revealing true relations.
Therefore, we are against any deployment without governance, informed consent, and human oversight aligned with ethical guidelines.

%% file: appendix_sections/additional_related_work.tex
\section{Extended Related work}
\label{appendix:additional_related_work}

\textbf{GNN recommender systems.}
Relational inductive biases are central to graph recommenders that represent user–item interactions as edges \citep{battaglia2018relational}.
From GCMC \citep{berg2017graph}, GNNs explicitly leverage higher-order connectivity,
including PinSage \citep{ying2018graph}, NGCF \citep{wang2019neural}, and simplified designs like LightGCN \citep{he2020lightgcn}.
Furthermore, knowledge-graph-aware models capture attribute/item relations \citep{wang2019kgat} and
complementary directions capture session and social structures \citep{wu2019session,fan2019graph}
or harness contrastive signals on graphs \citep{wu2021self, yu2023xsimgcl}.
These successes suggest that human choices and behaviors are inherently relational.
We draw on these insights to bridge two largely disjoint literature: LLM-based simulation and graph-based modeling.

\textbf{Discrete choice modeling.}
Classical `human simulation' in discrete choice has been based on latent variable frameworks \citep{train2009discrete, lin2023introduction},
random-utility models such as mixed logit \citep{mcfadden2000mixed},
latent-class/finite-mixture models \citep{greene2003latent},
integrated choice-and-latent-variable hybrids and hierarchical Bayes methods for individualized posteriors \citep{ben2002integration}.
More recently, graph learning offers a promise of complementary inductive bias that exploits interaction topology: \citet{tomlinson2024graph} outline several ways of integration, such as GNN-based chooser embeddings and propagation of local choice rates.
Building on these insights and the latent variable tradition, \ours casts discrete choice simulation as a link prediction on a heterogeneous graph.
By contrast, current LLM-based approaches primarily leverage parametric knowledge from pretraining.

\textbf{Text-attributed graphs (TAGs).}
TAGs integrate node and relation's text attributes with graph topology, letting models enjoy complementary signals.
Early work injected text features into matrix-factorization formulations or constructed word--document graphs \citep{yang2015network, yao2019graph}.
More recently, an LLM-to-GNN interplay has emerged:
(i) LLM as encoder/feature generator, using an LLM as an encoder whose embeddings serve as GNN inputs \citep{zhu2021textgnn};
(ii) alternating, EM-style training that decouples text and graph modules while co-training them via variational objectives \citep{zhao2022learning}; and
(iii) prompting LLMs to generate descriptions or explanations that enrich node attributes \citep{he2023harnessing}.
A complementary line of work conditions LLMs on graph structure through prompting and in-context learning, including AskGNN \citep{hu2024let} and GraphICL \citep{sun2025graphicl}.
Related efforts project graphs directly into an LLM’s token space or align GNN embeddings with token embeddings so that an LLM can reason over graph tokens \citep{chen2024llaga}.
In recommender systems, the inverse mapping uses LLM representations within learned graph / collaborative-filtering spaces or co-trains them with GNNs \citep{ren2024representation, sheng2025language}.
Collectively, these works underscore the complementarity of language and graph signals.
\ours leverages these insights for human simulation on discrete choice tasks,
especially on prediction for new questions where an appropriate mapping between language representations and graph representations is necessary.

%% file: appendix_sections/question_examples.tex
\section{Dataset Details}
\label{appendix:question_examples}

\subsection{OpinionQA}

\textsc{OpinionQA} \citep{santurkar2023whose} is a curated subset of the American Trends Panel (ATP) \citep{atp}.
It comprises 500 contentious questions drawn from 14 ATP survey waves, selected for large inter-group response differences.
For each anonymized participant, information across 9 demographic traits (age, gender, race or ethnicity, highest level of education, annual income, Census Bureau regions, religion, political affiliation, and political ideology) and their response to survey questions are available.
Survey items span a wide range of social topics, including race, politics, age-specific attitudes, media consumption, and views on the future of AI.
Owing to its breadth and diversity, \textsc{OpinionQA} has become a popular dataset for LLM-based human simulation or pluralistic preference alignment research \citep{hwang2023aligning, feng-etal-2024-modular, zhaogroup, moon2024virtual, kolluri2025finetuningllmshumanbehavior}.
Here we present three example questions from the dataset.

\begin{tcolorbox}[colback=white, colframe=black!75, 
  fonttitle=\bfseries, title=Question Example: OpinionQA (1), 
  boxrule=0.5pt, arc=2pt, outer arc=2pt]
\label{textbox:opinionqa_1}
Question: Which of the following would you say you prefer for getting news?

\bigskip
A. A print newspaper

B. Radio

C. Television

D. A social media site (such as Facebook, Twitter or Snapchat)

E. A news website or app
\end{tcolorbox}

\begin{tcolorbox}[colback=white, colframe=black!75, 
  fonttitle=\bfseries, title=Question Example: OpinionQA (2), 
  boxrule=0.5pt, arc=2pt, outer arc=2pt]
\label{textbox:opinionqa_2}

Question: In the future, what kind of an impact do you think the military will have in solving the biggest problems facing the country?

\bigskip
A. A very positive impact

B. A somewhat positive impact

C. A somewhat negative impact

D. A very negative impact
\end{tcolorbox}

\begin{tcolorbox}[colback=white, colframe=black!75, 
  fonttitle=\bfseries, title=Question Example: OpinionQA (3), 
  boxrule=0.5pt, arc=2pt, outer arc=2pt]
\label{textbox:opinionqa_3}
Question: For each, please indicate if you, personally, think it is acceptable. Casting an actor to play a character of a race or ethnicity other than their own

\bigskip
A. Always acceptable

B. Sometimes acceptable

C. Rarely acceptable

D. Never acceptable

E. Not sure
\end{tcolorbox}

\subsection{Twin-2K}

Twin-2K \citep{toubia2025twin} is a four-wave,
nationally representative U.S. panel fielded in January -- February 2025 on Prolific for LLM human simulation.
Each participant completed questions spanning demographic information, personality scales, cognitive ability tests, economic preference, heuristics-and-biases experiments, etc.
Among all questions from Twin-2K, we filtered for multiple-choice questions by removing short answer questions,
resulting in 150 questions total.
The authors release the full dataset publicly to support broader social-science research.

\begin{tcolorbox}[colback=white, colframe=black!75, 
  fonttitle=\bfseries, title=Question Example: Twin-2K (1), 
  boxrule=0.5pt, arc=2pt, outer arc=2pt]
\label{textbox:twin_2k_1}

Choose an option.

\bigskip
A. I don't feel like a failure

B. I feel that I have failed more than the average person

C. As I look back on my life, all I can see is a lot of failures

D. I feel I am a complete failure as a person
\end{tcolorbox}

\begin{tcolorbox}[colback=white, colframe=black!75, 
  fonttitle=\bfseries, title=Question Example: Twin-2K (3), 
  boxrule=0.5pt, arc=2pt, outer arc=2pt]
\label{textbox:twin_2k_3}

Antonym: Select the word that is most nearly the opposite in meaning to DEARTH

\bigskip
A. birth

B. brevity

C. abundance

D. splendor

E. renaissance
\end{tcolorbox}

\begin{tcolorbox}[colback=white, colframe=black!75, 
  fonttitle=\bfseries, title=Question Example: Twin-2K (2), 
  boxrule=0.5pt, arc=2pt, outer arc=2pt]
\label{textbox:twin_2k_2}

You have recently graduated from university, obtained a good job, and are buying a new car. A newly designed seatbelt has just become available that would save the lives of 95\% of the 500 drivers a year who are involved in a type of head-on-collision. (Approximately half of these fatalities involve drivers who were not at fault.) The newly designed seatbelt is not yet standard on most car models. However, it is available as a \$500 option for the car model that you are ordering. How likely is it that you would order your new car with this optional seatbelt?

\bigskip
A. very unlikely

B. unlikely

C. somewhat unlikely

D. somewhat likely

E. likely

F. very likely
\end{tcolorbox}

\subsection{Replication Study of the Dunning-Kruger Effect}

Replication study of the Dunning–Kruger effect (referred to as \textsc{Dunning-Kruger} throughout) \citep{jansen2021rational} is a single-wave, U.S.-based study.
3K participants first completed four pre-question survey items on self-expected problem-solving ability and confidence level,
then answered 20 multiple-choice questions on grammar and logical reasoning (five options per question, exactly one correct).
After the test, they completed four post-question survey items.
Our prediction target is the option a participant would select—not the correct answer—aligning with previous LLM human simulation objective \citep{binz2025foundation} and prediction tasks of interest in intelligent tutoring systems \citep{wang2020instructions}.
We present two samples from the 20 questions and one pre-question survey.

\begin{tcolorbox}[colback=white, colframe=black!75, 
  fonttitle=\bfseries, title=Question Example: Dunning-Kruger (1), 
  boxrule=0.5pt, arc=2pt, outer arc=2pt]
\label{textbox:dunning_kruger_1}

Compared to other participants in this study, how well do you think you will do?

Marking 90\% means you will do better than 90\% of participants,
marking 10\% means you will do better than only 10\%,
and marking 50\% means that you will perform better than half of the participants.

\end{tcolorbox}

\begin{tcolorbox}[colback=white, colframe=black!75, 
  fonttitle=\bfseries, title=Question Example: Dunning-Kruger (2), 
  boxrule=0.5pt, arc=2pt, outer arc=2pt]
\label{textbox:dunning_kruger_2}

Some part of the sentence is in square brackets.

Five choices for rephrasing that part follow the sentence; one choice repeats the original, and the other four are different.

Your task is to select the grammatically correct choice.

\bigskip
The school-age child faces a formidable task when during the first few years of classroom experiences [he or she is expected to master the printed form of language.]

\bigskip
A. he or she expects to master the printed form of language.

B. he or she is expected to master the printed form of language.

C. he or she faces expectations of mastering the printed form of language.

D. mastery of the printed form of language is expected of him or her.

E. mastery of print is expected by his or her teacher.

\end{tcolorbox}

\begin{tcolorbox}[colback=white, colframe=black!75, 
  fonttitle=\bfseries, title=Question Example: Dunning-Kruger (3), 
  boxrule=0.5pt, arc=2pt, outer arc=2pt]
\label{textbox:dunning_kruger_3}

Some part of the sentence is in square brackets.

Five choices for rephrasing that part follow the sentence; one choice repeats the original, and the other four are different.

Your task is to select the grammatically correct choice.

\bigskip
[The belief of ancient scientists was] that maggots are generated from decaying bodies and filth and are not formed by reproduction.

\bigskip
A. The belief of ancient scientists was

B. The ancient scientists beliefs were

C. The ancient scientists believe

D. The belief of ancient scientists were

E. The ancient belief of scientists was

\end{tcolorbox}

%% file: appendix_sections/graph_statistics.tex
\section{Graph Statistics}
\label{appendix:graph_statistics}

In this section, we present the graph statistics, including:
the number of nodes and edges of the graph formulation (\Cref{fig:figure_1}) and the construction of subgroup nodes.

\subsection{OpinionQA}

We followed the dataset filtering process of \cite{zhaogroup}.
Beginning with 76K participants in \textsc{OpinionQA} dataset,
filtering to those who answered at least 30 questions yields 19K individuals, 284 survey questions, and 695K (individual, question, choice) triples.
284 survey questions have total 1,103 choices, indicating that each survey question has 3.88 available choices on average.
Because the number of individual nodes (19K) is much larger than the number of choice nodes (1,103), choice nodes have much higher average degree than individual nodes.

To define subgroup nodes, we employ the 9 demographic attributes used in previous works \citep{santurkar2023whose}:
age, gender, race or ethnicity, education level, annual income, Census regions, religion, political affiliation, and political ideology.
This results in 48 subgroup nodes:

\begin{description}[leftmargin=5pt,labelindent=5pt,itemsep=3pt,parsep=0pt]
\item[\textbf{Age}]: 18-29, 30-49, 50-64, 65+
\item[\textbf{Race or ethnicity}]: White, Black, Hispanic, Asian, Other
\item[\textbf{Gender}]: Male, Female, Other
\item[\textbf{Education}]: Less than high school, High school graduate, Some college, no degree, Associate's degree, College graduate / some postgrad, Postgraduate
\item[\textbf{Annual income}]: Less than \$30,000, \$30,000 – \$50,000, \$50,000 – \$75,000, \$75,000 – \$100,000, \$100,000 or more
\item[\textbf{Region}]: Northeast, Midwest, South, West
\item[\textbf{Religion}]: Protestant, Roman Catholic, Mormon, Orthodox, Jewish, Muslim, Buddhist, Hindu, Atheist, Agnostic, Other, Nothing in particular
\item[\textbf{Political affiliation}]: Republican, Democrat, Independent, Something else
\item[\textbf{Political ideology}]: Very conservative, Conservative, Moderate, Liberal, Very liberal
\end{description}

We note that there can be different constructions of subgroup nodes,
either by considering additional individual features (e.g., marital status, country of birth, etc.)
or intersectional attributes as a single subgroup node (e.g., construct a subgroup node representing `age 18-29 male').
Future work can design their own subgroup nodes tailored to the specific need, and our construction is easily generalizable in those settings.

\subsection{Twin-2K}

\textsc{Twin-2K} dataset includes both multiple choice and short answer questions.
To align with our focus on discrete choice human simulation tasks, we exclude short-answer items, yielding 150 multiple-choice questions.
Because nearly all of the 2,000 participants responded to most multiple choice questions,
no individuals were removed by the minimum-30-responses criterion.
The dataset authors \citep{toubia2025twin} collect demographics using the same categories as \cite{santurkar2023whose}:
we reuse the identical 48 subgroup definitions as in \textsc{OpinionQA}.
The resulting graph contains 48 subgroup nodes, 2{,}000 individual nodes, 539 choice nodes, and 297K response edges.

\subsection{Replication Study of the Dunning-Kruger Effect}

Unlike the previous two datasets, this study does not include participant demographics.
Instead, we define individual features and derive subgroup nodes from responses to four pre-question survey items.
Choice nodes are constructed from 5 options for each of 20 questions, 100 in total.

\begin{description}[leftmargin=5pt,labelindent=5pt,itemsep=3pt,parsep=0pt]
\item[\textbf{Expected correctness}] (How many of the 20 questions do you think you will answer correctly?): 0, 1, 2, ..., 20, total 21 subgroups
\item[\textbf{Expected performance}] (How well do you think you will do in percentile?): 0-10, 11-20, 21-30, ..., 91-100, total 10 subgroups
\item[\textbf{Confidence of self}] (How difficult is recognizing correct grammar for you?): 0, 1, 2, ..., 10
\item[\textbf{Expected confidence of others}] (How difficult is recognizing correct grammar for the average participant?): 0, 1, 2, ..., 10, total 11 subgroups
\end{description}

%% file: appendix_sections/training_detail.tex
\section{Implementation Details}
\label{appendix:training_details}

This section details the implementation of the GNN (\Cref{main:architecture:details}) and the LLM baselines (\Cref{main:experiment:setup}).
We first present the general GNN training configuration in \Cref{appendix:training_details:config}, followed by the learnable input embedding tables in \Cref{appendix:training_details:table}.
Next, we instantiate the graph encoder in \eqref{eq:encoder} with three architectures—RGCN, GAT, and GraphSAGE—in \Cref{appendix:training_details:rgcn,appendix:training_details:gat,appendix:training_details:sage}, respectively.
\Cref{appendix:training_details:llm} describes the setup for the LLM-based methods.
Finally, \Cref{appendix:training_details:model_size} compares the model size of LLMs and \ours GNNs.

\subsection{GNN Training Configuration}
\label{appendix:training_details:config}

We implement GNNs based on PyTorch Geometric \citep{fey2019fast}. All trainable components of the GNN (learnable input embedding tables, graph encoder, and decoding temperature)
are optimized with AdamW optimizer \citep{loshchilov2017decoupled} using a learning rate of $5\times10^{-4}$ and weight decay of $10^{-3}$.
We use a cosine annealing learning rate scheduler and apply gradient clipping with a max norm of $0.1$.

Each GNN is trained for $1,000$ epochs with a patience of 30 (i.e., how many epochs the model would continue training after the validation loss stops decreasing).
In \Cref{main:advantages}, the reported GNN training time is measured from the beginning of the training until the termination by exceeding patience.
An epoch consists of $50n$ steps, where $n$ is the number of training graphs.
Concretely, each training graph is sampled $50$ times per epoch with independently re-drawn edge masks that split train response edges into message-passing edges and supervision edges.
This resampling reduces overfitting to a fixed edge partition and consistently improves validation accuracy and its variance.

\paragraph{Setting 3 (New questions).}
After the GNN is fully trained, we learn an LLM-to-GNN mapping as described in \Cref{eq:lproj}.
The mapping is obtained by solving ridge regression with a regularization strength $\alpha$.
Rather than a cross-validation, we choose $\alpha$ by directly maximizing validation prediction accuracy on held-out validation questions.
In \Cref{main:advantages}, the training time of LLM-to-GNN mapping is calculated as the time to extract LLM hidden states from textual features of choice nodes,
since solving the ridge regression takes negligible amount of time.
In practice, $\alpha \in [50, 500]$ performs best.

\subsection{Learnable Input Feature Table}
\label{appendix:training_details:table}

In \Cref{main:architecture:details},
we denote a learnable input feature table for subgroup nodes $\mathcal{S}$ as $Z_{\mathcal{S}} \in \mathbb{R}^{|\mathcal{S}|\times d_{\mathcal{S}}}$
and choice nodes $\mathcal{C}$ as $\smash{Z_{\mathcal{C}} \in \mathbb{R}^{|\mathcal{C}|\times d_{\mathcal{C}}}}$.
We set $d_{\mathcal{S}} = 16$ and $d_{\mathcal{C}} = 128$ for all settings on the \textsc{OpinionQA} dataset,
and $d_{\mathcal{S}} = 8$ and $d_{\mathcal{C}} = 64$ for all settings on the \textsc{Twin-2K} and \textsc{Dunning-Kruger} dataset.

\subsection{Relational Graph Convolution (RGCN)}
\label{appendix:training_details:rgcn}

We use a 2-layer RGCN \citep{schlichtkrull2018modeling}.
Following the feature table dimension in \ref{appendix:training_details:table},
input dimensions are $(16,1,128)$ for (subgroup, individual, choice) nodes on \textsc{OpinionQA} dataset
and $(8,1,64)$ on others. All hidden layers use the choice node's input dimension, i.e., 128 for \textsc{OpinionQA} and 64 for \textsc{Twin-2K} and \textsc{Dunning-Kruger}.

In \eqref{eq:encoder}, we present the general graph encoder forward pass as
\begin{equation}
\begin{aligned}
z_w^{(\ell+1)} \;=\; \sigma\!\Bigg(
\sum_{r\in\mathcal{R}}
\Big[
\operatorname*{AGG_{\mathit{r}}}_{v \in \mathcal{N}_r(w)}\;
\phi_{r}^{(\ell)}\!\big(z_w^{(\ell)},\, z_v^{(\ell)}\big)
\Big]
\;+\; \\
\phi_{\text{self}}^{(\ell)}\big(z_w^{(\ell)}\big)
\Bigg),\qquad \ell=0,\dots,L-1
\end{aligned}
\end{equation}
For RGCN, we use $\operatorname{ReLU}$ as the non-linear activation $\sigma$
and a mean pooling for $\operatorname{AGG}_{r}$ for all relations $r$.
Following the standard RGCN implementation, a relation-specific message passing is
\begin{equation}
\phi_{r}^{(\ell)}\!\big(z_w^{(\ell)},\, z_v^{(\ell)}\big)
= \frac{1}{|\mathcal{N}_r(w)|}\,\mathbf{W}^{(\ell)}_{r}\, z_v^{(\ell)},
\end{equation}
where the learnable $\mathbf{W}^{(\ell)}_{r}$ maps from the layer-$\ell$ embedding of the node $v$ to the layer-$(\ell{+}1)$ embedding dimension of node $w$;
the factor $|\mathcal{N}_r(w)|^{-1}$ provides degree normalization for relation $r$.
Similarly, self-loops use a learnable matrix $\mathbf{W}_{\text{self},t(w)}^{(\ell)}$ that is node-type specific:
\begin{equation}
\phi_{\text{self}}^{(\ell)}\big(z_w^{(\ell)}\big)
= \mathbf{W}_{\text{self},t(w)}^{(\ell)}\, z_w^{(\ell)}.
\end{equation}
where $t(w)$ represents the node type of the node $w$.
Additionally, we apply a post-activation LayerNorm \citep{ba2016layer} and dropout with rate $0.5$ at all layers of the graph encoder.

\subsection{Graph Attention Network (GAT)}
\label{appendix:training_details:gat}

\Cref{eq:encoder} is implemented with a multi-head Graph Attention Network \citep{velivckovic2017graph}
\begin{equation}
\label{eq:gat:update}
\begin{aligned}
z_w^{(\ell+1)}
&= \sigma\!\left(
\sum_{r\in\mathcal{R}}
\Bigg[
\mathop{\Big\|}\limits_{h=1}^{H_\ell}\;
\sum_{v \in \mathcal{N}_r(w)\cup\{w\}}
\right. \\
&\qquad\left.
\alpha_{wv,r}^{(\ell,h)}\;
\boldsymbol{\Theta}_{t,r}^{(\ell,h)}\, z_v^{(\ell)}
\Bigg]
\right)
\end{aligned}
\end{equation}
where $\|$ denotes concatenation across heads,
$h$ indicates the head index ranging from 1 to the number of heads in the $\ell$-th layer ($H_\ell$),
and the attention coefficient $\alpha$ for the layer-$\ell$ head-$h$ relation-$r$ from the source node $v$ to the target node $w$ is
\begin{equation}
\label{eq:gat:alpha}
\begin{aligned}
&
\alpha_{wv,r}^{(\ell,h)}
=
\operatorname*{softmax}_{v \in \mathcal{N}_r(w)\cup\{w\}}
\Big(
\operatorname{LeakyReLU}\Big( \\
&
\mathbf{a}_{s,r}^{(\ell,h)\top}\,\boldsymbol{\Theta}_{s,r}^{(\ell,h)} z_w^{(\ell)}
+
\mathbf{a}_{t,r}^{(\ell,h)\top}\,\boldsymbol{\Theta}_{t,r}^{(\ell,h)} z_v^{(\ell)}
\Big)
\Big).
\end{aligned}
\end{equation}
where $\mathbf{a}^{(\ell,h)}_{s,r}$ and $\mathbf{a}^{(\ell,h)}_{t,r}$ are learnable source and target scoring vectors,
$\mathbf{\Theta}^{(\ell,h)}_{s,r}$ and $\mathbf{\Theta}^{(\ell,h)}_{t,r}$ are learnable source and target feature transformation matrix,
and $\operatorname{LeakyReLU}$ is a LeakyReLU function with a negative slope of 0.2 as in the default implementation of PyTorch Geometric.
Softmax is performed over all neighboring nodes of $w$ defined by the relation $r$ and $w$ itself.

We use a 2-layer GAT.
Following the input table dimension in \ref{appendix:training_details:table},
input feature dimensions are $(16,1,128)$ for (subgroup, individual, choice) nodes on the \textsc{OpinionQA} dataset and $(8,1,64)$ on others.
All hidden layers use the choice node's input dimension (128 for \textsc{OpinionQA}, 64 for \textsc{Twin-2K} and \textsc{Dunning-Kruger})
with 4 heads in the first layer (per-head size of 32 or 16) and 1 head in the second layer (per-head size of 128 or 64), keeping the hidden size unchanged across layers.

We set $\sigma=\operatorname{ReLU}$, and apply post-activation LayerNorm \citep{ba2016layer}.
We also apply dropout at rate $0.4$ to the normalized attention coefficients $\alpha$ and at rate $0.5$ to the post-activation node embeddings between layers.

\subsection{GraphSAGE}
\label{appendix:training_details:sage}

We instantiate the generic graph encoder in \eqref{eq:encoder} with a GraphSAGE operator \citep{hamilton2017inductive}.
For each relation $r\in\mathcal{R}$ and given a target node $w$, we first compute a relation-specific mean-pooled neighbor message
\begin{equation}
m_{w,r}^{(\ell)}
\;=\;
\operatorname*{MEAN}_{v \in \mathcal{N}_r(w)}
\big(\,\mathbf{\Theta}_{r}^{(\ell)}\, z_v^{(\ell)}\big),
\label{eq:sage:meanmsg}
\end{equation}
where $\mathbf{\Theta}_{r}^{(\ell)}$ is a learnable matrix that maps the layer-$\ell$ embedding of a source node $v$ to the layer-$(\ell{+}1)$ embedding space of the target node for relation $r$.
Messages from all relations are summed and combined with a learnable root (self) transformation $\smash{\mathbf{\Theta}_{\text{self}}^{(\ell)}}$.
Subsequently, the embedding is $L2-$normalized and passed through a non-linear activation:
\begin{equation}
\label{eq:sage:update}
\begin{aligned}
z_w^{(\ell+1)}
&= \sigma\!\Big(
\operatorname{Normalize}\Big( \\
&\qquad \boldsymbol{\Theta}_{\text{self}}^{(\ell)}\, z_w^{(\ell)}
+ \sum_{r\in\mathcal{R}} m_{w,r}^{(\ell)}
\Big)
\Big).
\end{aligned}
\end{equation}
We set $\sigma=\operatorname{ReLU}$, apply post-activation LayerNorm \citep{ba2016layer} at every layer, and use dropout with rate $0.5$ on the post-activation node embeddings between layers.

We use a 2-layer GraphSAGE.
Following the input feature dimensions in \Cref{appendix:training_details:table},
input sizes are $(16,1,128)$ for (subgroup, individual, choice) nodes on \textsc{OpinionQA} and $(8,1,64)$ on Twin-2K.
All hidden layers use the choice-node width, i.e., $128$ for \textsc{OpinionQA} and $64$ for Twin-2K.

\subsection{LLM}
\label{appendix:training_details:llm}

For all LLM prompting experiments, we used 2$\times$NVIDIA A100 80GB (SXM4) and vLLM framework \citep{kwon2023efficient}.
For selecting in-context examples in few-shot prompting and Agentic CoT,
we encode each question text with \texttt{gemini-embedding-001} embedding model and compute cosine similarities between the target question and candidate in-context example questions.
Following \citet{hwang2023aligning}, the selected examples are ordered by ascending cosine similarity, from least to most similar.
To ensure consistent information access across methods, in-context examples are drawn exclusively from the training set and not from the validation set.

For all LLM fine-tuning methods, we also used 2$\times$NVIDIA A100~80GB (SXM4) and built on Llama-cookbook codebase.
Each run trained for three epochs using the model’s default precision,
and we selected the checkpoint with the lowest validation loss.
We largely followed hyperparameter setting of \citet{suh2025language},
tuning the learning rate over \{1e-4, 2e-4, 4e-4\} and settled on 2e-4.
Training used LoRA with rank~8, $\alpha\!=\!32$, and dropout~0.05, applied to the attention query and value matrices, following the implementation details of main experiment in the original LoRA paper \citep{hu2022lora}.
We optimized with Adam optimizer \citep{kingma2014adam} and the effective batch size was fixed to 256
by setting per-GPU batches and gradient accumulation steps to fit GPU VRAM.
For ablation experiments with higher LoRA rank (rank 256) or full fine-tuning, refer to \Cref{appendix:ablations}.

\subsection{Model Parameters}
\label{appendix:training_details:model_size}

In \Cref{tab:parameters}, we report parameter counts for \ours and the LLMs, following the implementation details in the previous sections.
For LLMs fine‑tuned with LoRA, the trainable parameter count equals the number of LoRA adapter parameters, much smaller than the total parameter count.
Because both training and inference still require loading the full model weights, we use the total parameter count when comparing model size.
The size of the GNN (\ours) varies between datasets because we select the hidden dimension per dataset, as noted above.

\input{table/table_parameters}

%% file: table/table_parameters.tex
\begin{table*}[ht]
\caption{
Number of parameters for each model.
K, M, and B stand for $10^3$, $10^6$, $10^9$, respectively.
}
\centering
\label{tab:parameters}
\resizebox{0.8\textwidth}{!}{
\begin{tabular}{c|ccc|ccc}
\toprule
\textbf{\# parameters} & \multicolumn{3}{c}{LLM} & \multicolumn{3}{c}{\ours (RGCN)} \\
\midrule
Model & LLaMA2-7B & Mistral-7B & Qwen3-8B & \multicolumn{3}{c}{-} \\
\midrule
Dataset & \multicolumn{3}{c|}{-} & \textsc{OpinionQA} & \textsc{Twin-2k} & \textsc{Dunning-Kruger} \\
\midrule
Trainable & 4.19 M & 3.41 M & 3.83 M & 420 K & 110 K & 110 K \\
Total     & 6.61 B & 7.24 B & 8.19 B & 420 K & 110 K & 110 K \\
\bottomrule
\end{tabular}
}
\end{table*}

%% file: appendix_sections/addtional_experiment.tex
\section{Additional Experiment Results}
\label{appendix:additional_experiment}

\subsection{Extended Figure of Compute Hours}

\begin{figure*}[t]
    \centering
    \includegraphics[width=\textwidth]{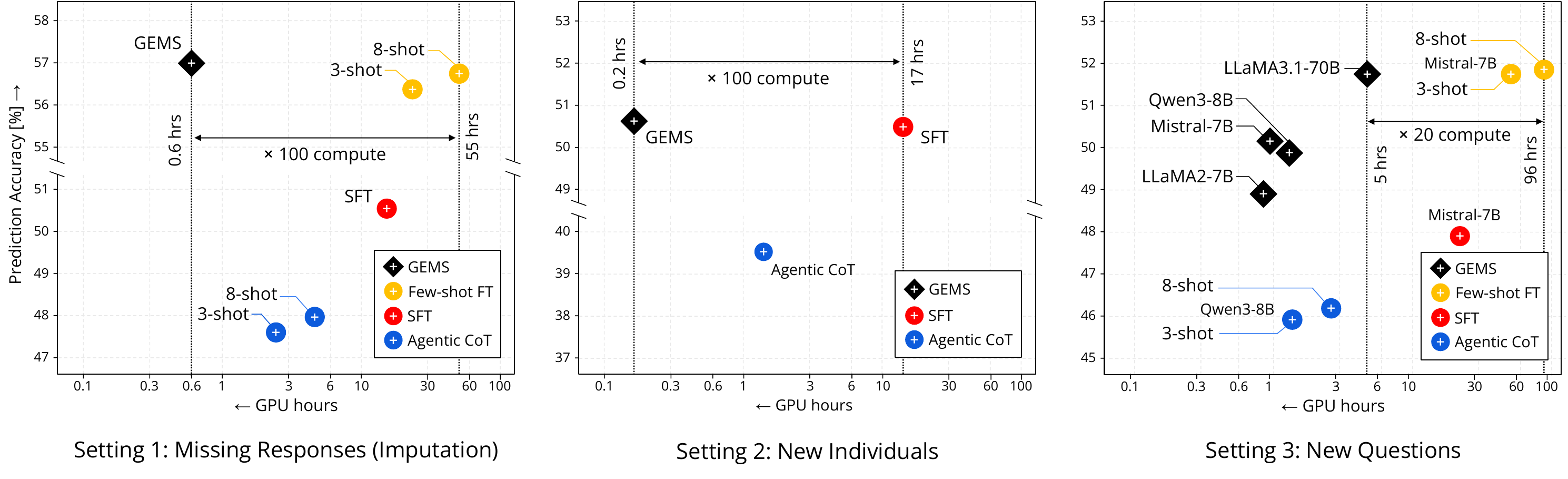}
    \caption{
    Accuracy \textit{vs.} GPU-hours (A100-80GB-SXM4) on the \textsc{OpinionQA} dataset by task setting and method.
    Zero-/few-shot prompting accuracies fall below the plotted y-range.
    For LLM-based methods, we report the best result across three LLMs (LLaMA-2-7B, Mistral-7B-v0.1, and Qwen3-8B).
    For \ours, we report the best result across three models (RGCN, GAT, and SAGE) for setting 1 \& 2,
    and report across different LLMs for setting 3.
    }
    \label{fig:figure_performance_compute_full}
\end{figure*}

\Cref{fig:figure_performance_compute_full} shows the prediction accuracy and compute hours of different LLM-based methods and \ours. Across all settings, \ours requires significantly less compute compared to LLM methods of comparable performance, and strictly outperforms LLM methods of comparable compute hours.

\subsection{Embedding visualization}
\label{appendix:additional_experiment:embedding_visualization}

\begin{figure*}[t]
    \centering
    \includegraphics[width=0.8\linewidth]{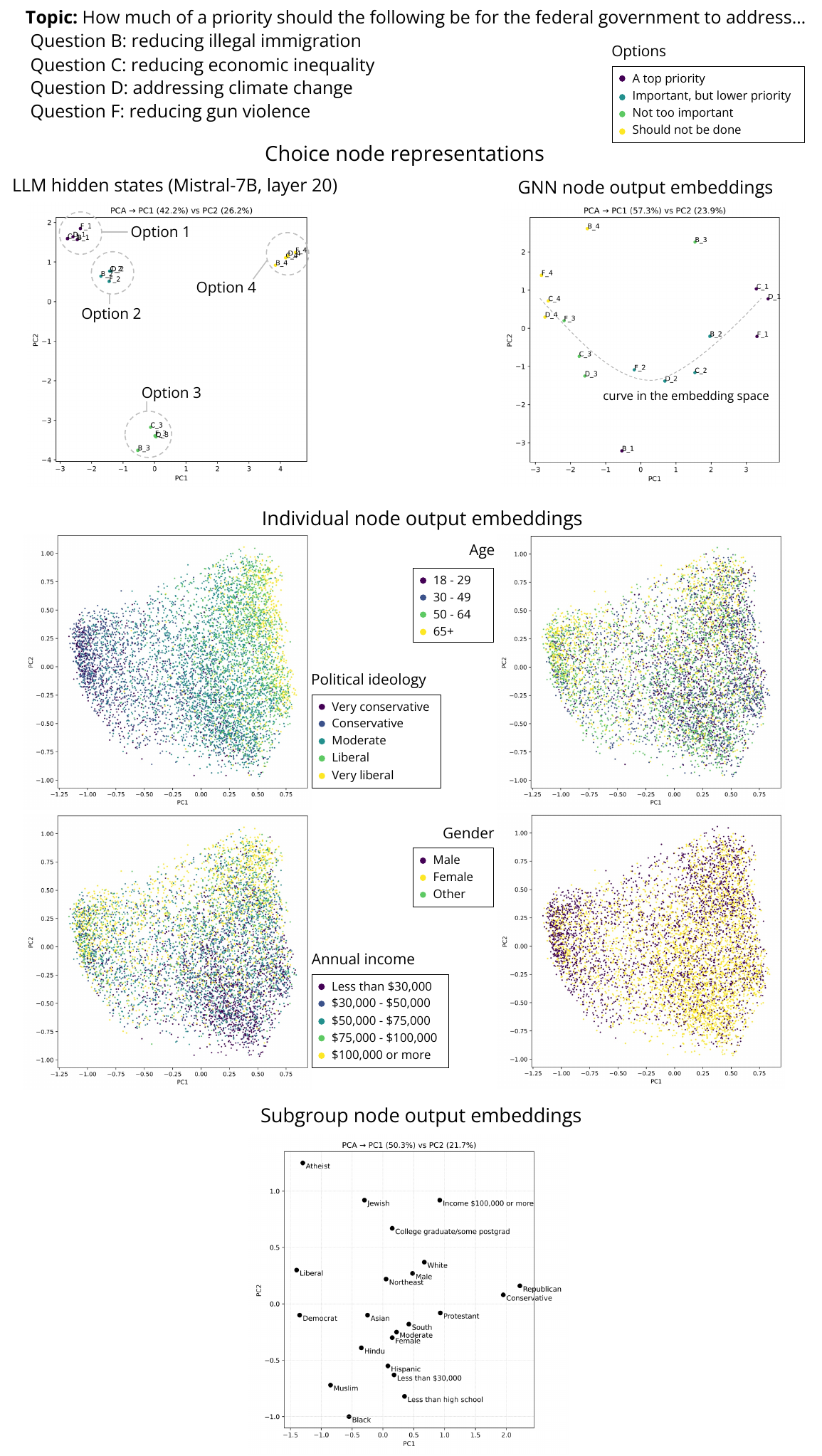}
    \caption{
    Visualization of LLM hidden states and GNN node embeddings on the first and second components of principal component analysis.
    }
    \label{fig:figure_embeddings}
\end{figure*}

\Cref{fig:figure_embeddings} visualizes LLM hidden states and GNN embeddings for four example questions in \textsc{OpinionQA} dataset.
Each question asks how high a priority the federal government should give to an issue:
(B) reducing illegal immigration,
(C) reducing economic inequality,
(D) addressing climate change, and
(F) reducing gun violence,
with four response options ranging from `top priority' to `should not be done.'
This results in 16 choice nodes in total. All plots show the first two principal components of principal component analysis (PCA).

\textbf{Embedding structure of choice nodes.}
The top left panel plots the LLM hidden states for the 16 choice nodes:
points cluster by option, producing four clusters one per option but not clearly indicating what semantic meaning each choice has.
The top right panel shows choice nodes' output node embeddings after the first training stage described in \Cref{main:experiment:setting3}.
Here, choices for three questions (C, D, F) are located along a common one-dimensional trajectory in the PCA plane,
whereas the choices for question B align along a distinct trajectory.
From this observation, we can infer that three questions (C, D, F) are closely related while one question (B) sits on a slightly different social issue dimension,
which is consistent with prior observation from survey researchers \citep{pew2024issues2024election}.

\textbf{Embedding structure of individual nodes.}
The remaining panels plot GNN output node embeddings of individual nodes,
with colors indicating an individual feature per panel (annual income, political ideology, age, or gender).
The PCA axes exhibit interpretable variation: PC1 aligns most strongly with political ideology feature and PC2 with income.
Yet, points within any given subgroup remain dispersed, indicating substantial within-group heterogeneity.
We note that the prediction is made by taking dot-product between each individual node embedding and the four choice node embeddings, followed by the softmax for multinomial distribution over options.

\subsection{Prediction with Additional LLMs}
\label{appendix:additional_experiment:different_llms}

\begin{table*}[!t]
    \centering
    \small
    \setlength{\tabcolsep}{3.5pt}
    \caption{
        Extended evaluation with open-weight LLMs on the \textsc{OpinionQA} dataset, Setting~1 (missing responses).
        $k$ indicates the number of few-shot examples.
        Results for LLaMA-2-7B, Mistral-7B-v0.1, and Qwen3-8B are identical to those in Table~\ref{table:table_transductive_individual}.
    }
    \begin{tabular}{ll cccc cc cc c}
    \toprule
    & & \multicolumn{4}{c}{\textbf{LLaMA}} & \multicolumn{2}{c}{\textbf{Mistral}} & \multicolumn{2}{c}{\textbf{Qwen3}} & \textbf{GPT} \\
    \cmidrule(lr){3-6} \cmidrule(lr){7-8} \cmidrule(lr){9-10} \cmidrule(lr){11-11}
    \textbf{Methods} & $k$ & 2-7B & 2-70B & 3.1-8B & 3.1-70B & 7B-v0.1 & 24B-2501 & 8B & 32B & OSS-20B \\
    \midrule
    Random & & \multicolumn{9}{c}{27.87} \\
    \midrule
    Zero-shot & & 29.18 & 36.47 & 38.71 & 43.45 & 34.63 & 41.79 & 39.38 & 40.42 & 35.71 \\
    \midrule
    \multirow{4}{*}{Few-shot}
    & 3  & 38.54 & 40.76 & 44.04 & 46.04 & 42.52 & 45.85 & 42.21 & 43.53 & 42.18 \\
    & 5  & 39.41 & 43.26 & 44.22 & 47.35 & 44.43 & 45.82 & 42.69 & 46.04 & 44.53 \\
    & 8  & 37.91 & 40.34 & 42.26 & 44.55 & 45.78 & 45.27 & 43.66 & 44.28 & 41.47 \\
    & 13 & 37.78 & 42.45 & 43.24 & 49.66 & 44.03 & 46.41 & 44.03 & 46.72 & 42.50 \\
    \midrule
    \multirow{4}{*}{\shortstack[l]{Agentic\\CoT}}
    & 3  & 32.19 & 43.66 & 42.80 & 49.32 & 41.37 & 46.01 & 47.63 & 47.57 & 44.42 \\
    & 5  & 30.72 & 45.82 & 42.96 & 49.55 & 38.92 & 48.22 & 47.92 & 48.40 & 45.90 \\
    & 8  & 28.80 & 42.63 & 42.15 & 47.72 & 38.43 & 46.24 & 47.97 & 48.44 & 41.94 \\
    & 13 & 28.05 & 45.34 & 42.81 & 48.06 & 38.37 & 48.67 & 48.88 & 50.70 & 44.03 \\
    \midrule
    SFT          &   & 49.41 & -- & -- & -- & 50.56 & -- & 48.84 & -- & -- \\
    Few-shot FT  & 3 & 55.59 & -- & -- & -- & 56.31 & -- & 55.09 & -- & -- \\
    Few-shot FT  & 8 & 55.98 & -- & -- & -- & \textbf{56.76} & -- & 55.61 & -- & -- \\
    \bottomrule
    \end{tabular}
    \label{tab:extended_open}
\end{table*}

\begin{table*}[!t]
\centering
\small
\caption{
    Evaluation of frontier proprietary models on the \textsc{OpinionQA} dataset, Setting~1 (missing responses), with prompting methods.
    $k$ indicates the number of few-shot examples.
    Two bottom rows represent the best LLM fine-tuning and GEMS results from Table~\ref{table:table_transductive_individual} for comparison.
}
\begin{tabular}{ll cccc}
\toprule
\textbf{Methods} & $k$ & GPT-5-Mini & GPT-5.2 & Claude Sonnet 4.5 & Gemini 3 Flash \\
\midrule
Random & & \multicolumn{4}{c}{27.87} \\
\midrule
Zero-shot & & 45.06 & 45.35 & 45.23 & 46.06 \\
\midrule
\multirow{2}{*}{Few-shot}
& 3 & 46.99 & 48.16 & 47.27 & 48.25 \\
& 8 & 47.52 & 49.72 & 50.63 & 50.49 \\
\midrule
\multirow{2}{*}{\shortstack[l]{Agentic CoT}}
& 3 & 48.83 & 48.68 & 49.34 & 49.19 \\
& 8 & 50.17 & 52.01 & 51.84 & \textbf{52.78} \\
\midrule
\midrule
\multicolumn{2}{l}{Few-shot FT (best)} & \multicolumn{4}{c}{56.76} \\
\multicolumn{2}{l}{GEMS (best)} & \multicolumn{4}{c}{\textbf{57.00}} \\
\bottomrule
\end{tabular}
\label{tab:extended_proprietary}
\end{table*}

We expand the evaluation on Setting~1 (predicting missing responses) of the \textsc{OpinionQA} dataset to include both
open-weight models (LLaMA3.1-8B, LLaMA3.1-70B, Mistral-Small-24B-2501, Qwen3-32B, GPT-OSS-20B \citep{dubey2024llama,yang2025qwen3,openai2025gptoss120bgptoss20bmodel})
and frontier proprietary models (GPT-5-Mini, GPT-5.2, Claude Sonnet~4.5, and Gemini~3~Flash).
 
\textbf{Open-weight models.}
Table~\ref{tab:extended_open} reports results for open-weight models of varying sizes.
Consistent with Tables~\ref{table:table_transductive_individual}--\ref{table:table_inductive_question},
larger and more recent models generally perform better, with the largest gains appearing under Agentic CoT where reasoning ability is most critical.
This trend is most pronounced for Qwen3, a reasoning model family.
However, even the best prompting result among open-weight models (Qwen3-32B, Agentic CoT with 13 examples, 50.70\%) falls short of few-shot fine-tuning (56.76\%) and GEMS (57.00\%).

\textbf{Frontier proprietary models.}
Table~\ref{tab:extended_proprietary} extends the comparison to frontier proprietary models, evaluated with promting methods (as fine-tuning is unavailable).
First, frontier models achieve notable gains over smaller open-weight models.
For example, with 8-shot Agentic CoT, Gemini~3~Flash reaches 52.78\% and GPT-5.2 reaches 52.01\%, compared to 47.97\% for Qwen3-8B and 50.70\% for Qwen3-32B.
Second, even the best prompting results from frontier models do not surpass fine-tuning methods or GEMS:
the strongest prompting result (Gemini~3~Flash, Agentic CoT with $k{=}8$: 52.78\%) falls short of few-shot fine-tuning (56.76\%) and GEMS (57.00\%).
This indicates diminishing returns from scaling LLMs in prompting for close-ended simulation tasks, and confirms that GEMS remains competitive even against state-of-the-art proprietary models.

%% file: appendix_sections/ablations.tex
\clearpage
\newpage
\section{Ablations}
\label{appendix:ablations}

\subsection{LLM baselines: LoRA fine-tuning with larger ranks, and full fine-tuning}

\input{table/ablation_different_ranks}

We perform additional LLM fine-tuning baseline experiments (SFT, few-shot FT) and
show that higher LoRA ranks or full fine-tuning do not necessarily improve performance over the main results (\Cref{table:table_transductive_individual,table:table_inductive_individual,table:table_inductive_question}) with LoRA rank 8.
\Cref{table:ablation_different_ranks} shows the test prediction accuracy of Mistral-7B on \textsc{Dunning-Kruger} dataset for a single seed.
Across all three settings, and in each setting across different numbers of few-shot examples (except for setting 2, new individuals, where past responses of new individuals are not available)
our main result with LoRA $r=8$ does not necessarily underperform fine-tuning runs with larger capacity.

\subsection{Effect of hidden states across layers}

\begin{figure}[h]
    \centering
    \includegraphics[width=\linewidth]{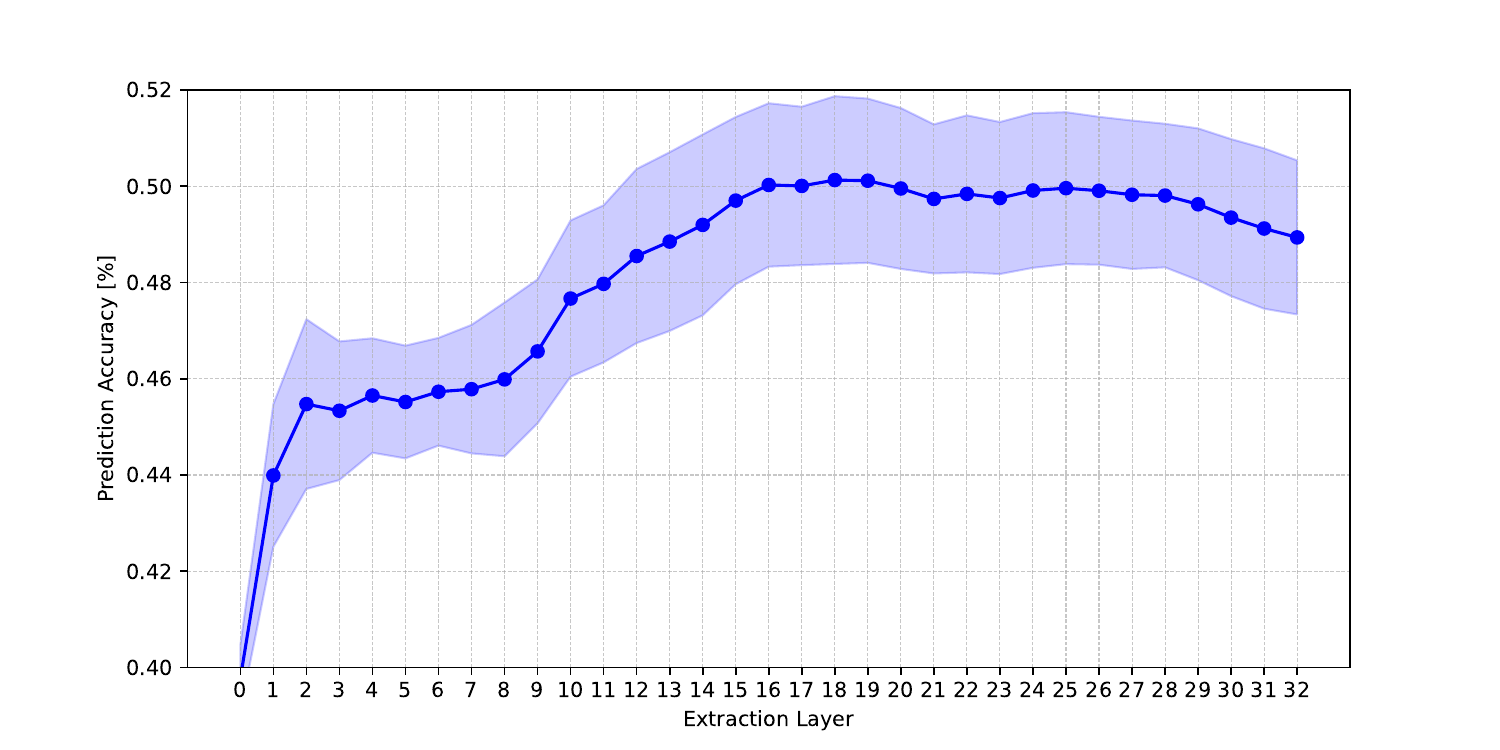}
    \caption{
    Mean and standard deviation of prediction accuracy on setting~3 (new questions) of \textsc{OpinionQA} dataset when extracting hidden states from different layers of Mistral-7B-v0.1 (\Cref{table:table_inductive_question}).
    Layer~0 is the post-embedding activation and layer~32 is the final pre-LM head activation.
    }
    \label{fig:figure_prediction_layer}
\end{figure}

\Cref{fig:figure_prediction_layer} shows \ours accuracy on the \textsc{OpinionQA} dataset with different layers of LLM (Mistral-7B-v0.1) to extract the hidden state from.
In practice, we choose the layer that maximizes accuracy on validation questions, which turned out to be layer-18 for Mistral-7B-v0.1 on the \textsc{OpinionQA} dataset.
We note that the best transformer layer changes depending on the LLM and dataset, and has to be found through search with prediction accuracy on validation questions.
Consistent with prior works on probing and interpretability \citep{kimlinear, tigges2023linear},
middle-to-late layers generally provide the most semantically useful and transferable language representations.

\subsection{Effect of the number of LLM--GNN representation pairs}

\begin{figure}[h]
    \centering
    \includegraphics[width=\linewidth]{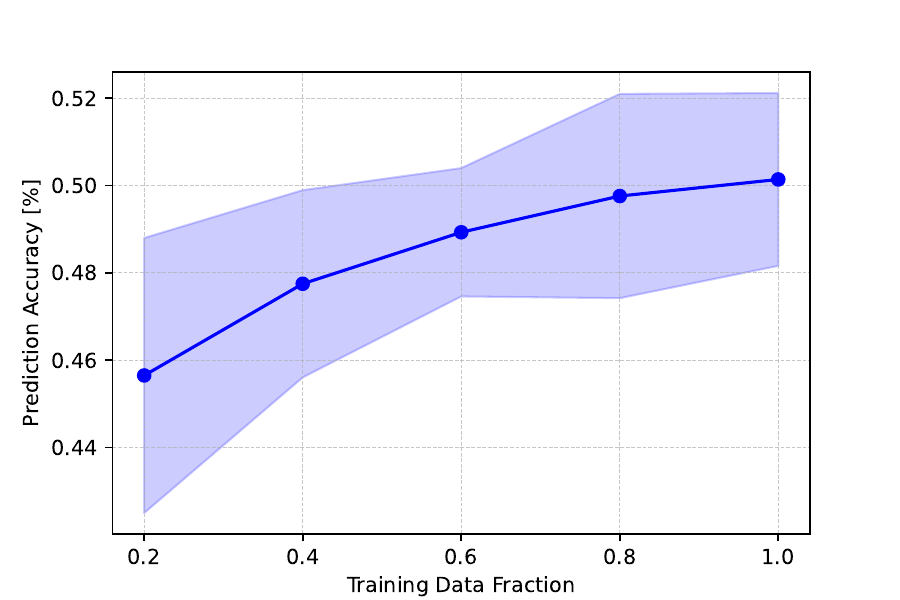}
    \caption{
    Mean and standard deviation of prediction accuracy on Setting~3 (new questions) of the \textsc{OpinionQA} dataset using hidden states from layer-18 of Mistral-7B-v0.1.
    The $x$-axis denotes the fraction of choice nodes in the training graph used to fit the LLM-to-GNN projection in \Cref{eq:lproj}.
    Accuracy improves as more paired examples are used, indicating that sufficient supervision is required to learn a map from LLM hidden states to the GNN output embedding space.
    }
    \label{fig:figure_training_data_fraction}
\end{figure}

Learning the LLM-to-GNN representation projection with linear mapping requires paired examples of an LLM hidden state and its corresponding GNN output node embedding.
Because this mapping lacks a linguistic prior, performance may degrade sharply when trained on too few pairs.
We validate this with an ablation that fits the mapping using only 20\%, 40\%, 60\%, and 80\% of the available pairs (fractions taken over choice nodes $\mathcal{C_{\text{train}}}$ in \Cref{eq:lproj}) and evaluate on the new question setting.
As shown in \Cref{fig:figure_training_data_fraction}, reducing the number of pairs leads to a rapid drop in accuracy, showing the sample size sensitivity of the mapping.

%% file: table/ablation_different_ranks.tex
\begin{table*}[t]
  \centering
  \caption{
  LLM fine-tuning performance with different trainable parameter size.
  Base LLM is Mistral-7B-v0.1, and fine-tuning performed on \textsc{Dunning-Kruger} dataset for all three settings.
  Setting 1, 2, 3 are predicting missing responses (imputation), responses of new individuals, and responses for new questions, respectively.
  }
  \label{table:ablation_different_ranks}
    \resizebox{0.65\textwidth}{!}{  
    \begin{tabular}{c|ccc|c|ccc}
    \toprule
    \textbf{Setting} & 
    \multicolumn{3}{c|}{\textbf{Setting 1}} & 
    \multicolumn{1}{c|}{\textbf{Setting 2}} & 
    \multicolumn{3}{c}{\textbf{Setting 3}} \\
    \midrule
    \textbf{Few-shots} & 0 & 3 & 8 & - & 0 & 3 & 8 \\
    \midrule
    LoRA, $r=8$      & 56.35 & 56.85 & 56.92 & 56.54 & 41.28 & 46.34 & 40.48 \\
    LoRA, $r=256$    & 56.35 & 56.61 & 57.38 & 56.68 & 42.75 & 46.27 & 42.33 \\
    Full fine-tuning & 56.35 & 56.68 & 57.36 & 56.58 & 42.78 & 46.25 & 41.85 \\
    \bottomrule
    \end{tabular}
    }
\end{table*}

%% file: appendix_sections/prompts.tex
\section{Prompts to LLM}
\label{appendix:prompts_llm}

\begin{description}[leftmargin=3pt,labelindent=3pt,itemsep=5pt,parsep=0pt]
    \item[Zero-shot prompt:]
    Provide an individual's features (demographics for \textsc{OpinionQA} and \textsc{Twin-2K}, pre-question survey responses for \textsc{Dunning-Kruger} dataset) in text form, followed by the question.
    Details about the list of individual features are defined in \Cref{appendix:graph_statistics}.
    When an individual feature is missing (e.g., age is unknown), we omit it in the prompt rather than explicitly stating its absence (e.g., ``Age: unknown'').
    Zero-shot prompts are used in zero-shot prompting, and supervised fine-tuning (SFT).
    
    \item[Few-shot prompt] (with variable $k$ in-context examples):
    Provide the individual's features, followed by $k$ prior responses to related questions (see \Cref{appendix:ablations} for how we select related questions).
    Append the target question at the end.
    Few-shot prompts are used in few-shot prompting, and few-shot fine-tuning (few-shot FT).
    
    \item[Agentic CoT prompt:]
    We directly adopt from \cite{park2024generative} with minimal modifications.
    The method consists of two stages.
    First, the individual's features and prior responses are given to an \emph{expert reflection} module, which produces concise observations about the person's stances.
    Second, these observations, together with the individual's context, are passed to a prediction module that outputs an answer in the JSON format.
\end{description}

All examples presented in this section use synthetic profiles and responses, not real individuals, to protect privacy (\Cref{appendix:limitations_potential_risks}).
For fine-tuning, we apply cross-entropy loss to the single answer token immediately following the input prompt.
We note that Qwen-3 \citep{yang2025qwen3} and GPT-OSS \citep{openai2025gptoss120bgptoss20bmodel} use distinct response formats and detail the required tokenization and formatting in \Cref{appendix:note_gpt_oss}.

\begin{tcolorbox}[breakable, colback=white, colframe=black!75, 
  fonttitle=\bfseries, title=Prompt Example: Zero-shot, 
  boxrule=0.5pt, arc=2pt, outer arc=2pt]
\label{textbox:task_1}
\textbf{System}

Respond to the following question by choosing one of the available options,
and strictly answering with the option letter (e.g., 'A', 'B', etc.).
Do not provide any additional text or explanation.

\bigskip
\textbf{User}

Answer the following question as if your personal information is as follows:

Personal identification number: 12345.0

Age: 50-64

Race or ethnicity: White

Gender: Female

Education level: Some college, no degree

Income level: less than \$30,000

Region of residence: West

Religion: Nothing in particular

Political party affiliation: Independent

Political ideology: Moderate

\bigskip
Question: Thinking about the nation's economy,
how would you rate economic conditions in this country today?

A. Excellent

B. Good

C. Only fair

D. Poor

\bigskip
Answer:

\end{tcolorbox}

\begin{tcolorbox}[breakable, colback=white, colframe=black!75, 
  fonttitle=\bfseries, title={Prompt Example: Few-shot ($k=2$)}, 
  boxrule=0.5pt, arc=2pt, outer arc=2pt]
\label{textbox:task_2}

\textbf{System}

Respond to the following question by choosing one of the available options,
and strictly answering with the option letter (e.g., 'A', 'B', etc.).
Do not provide any additional text or explanation.

\bigskip
\textbf{User}

Answer the following question as if your personal information is as follows:

Personal identification number: 12345.0

Age: 50-64

Race or ethnicity: White

Gender: Female

Education level: Some college, no degree

Income level: less than \$30,000

Region of residence: West

Religion: Nothing in particular

Political party affiliation: Independent

Political ideology: Moderate

\bigskip
Question: How much, if at all, do you think the following proposals would do to reduce economic inequality in the U.S.? Expanding government benefits for the poor

A. A great deal

B. A fair amount

C. Not too much

D. Nothing at all

Answer:

\bigskip
\textbf{Assistant}

A. A great deal

\bigskip
\textbf{User}

Question: How much, if at all, do you think some people work harder than others contributes to economic inequality in this country?

A. Contributes a great deal

B. Contributes a fair amount

C. Contributes not too much

D. Contributes not at all

Answer:

\bigskip
\textbf{Assistant}

A. Contributes a great deal

\bigskip
\textbf{User}

Question: Thinking about the nation's economy,
how would you rate economic conditions in this country today?

A. Excellent

B. Good

C. Only fair

D. Poor

\bigskip
Answer:

\end{tcolorbox}

\begin{tcolorbox}[breakable, colback=white, colframe=black!75, 
  fonttitle=\bfseries, title={Prompt Example: Agentic CoT \\ Reflection module ($k=2$)}, 
  boxrule=0.5pt, arc=2pt, outer arc=2pt]
\label{textbox:genagent}

\textbf{User}

[Participant's information]

Age: 50-64

Race or ethnicity: White

Gender: Female

Education level: Some college, no degree

Income level: less than \$30,000

Region of residence: West

Religion: Nothing in particular

Political party affiliation: Independent

Political ideology: Moderate

\bigskip
[Participant's prior responses]

Question: How much, if at all, do you think the following proposals would do to reduce economic inequality in the U.S.? Expanding government benefits for the poor

A. A great deal

B. A fair amount

C. Not too much

D. Nothing at all

Answer: A. A great deal

\bigskip
Question: How much, if at all, do you think some people work harder than others contributes to economic inequality in this country?

A. Contributes a great deal

B. Contributes a fair amount

C. Contributes not too much

D. Contributes not at all

Answer: A. Contributes a great deal

\bigskip
Imagine you are an expert social scientist (with a PhD) taking notes while observing this content. Write observations/reflections about the person's stances about key societal issues. (You should make more than 5 observations and fewer than 20. Choose the number that makes sense given the depth of the content above.)

\end{tcolorbox}

\begin{tcolorbox}[breakable, colback=white, colframe=black!75, 
  fonttitle=\bfseries, title={Prompt Example: Agentic CoT \\ Reflection module output (truncated)}, 
  boxrule=0.5pt, arc=2pt, outer arc=2pt]
\label{textbox:genagent_expert_output}

\textbf{Assistant}

**Observations \& Reflections on the Participant’s Societal Stances**

\bigskip
1. **Political identity**  

   - Identifying as an *independent* with a *moderate* ideology indicates a willingness to cross party lines on key issues.  
   
   - Such a position often correlates with a preference for incremental, evidence‑based reforms rather than ideological extremes.

\bigskip
2. **Attitude towards economic inequality**

...

\end{tcolorbox}

\begin{tcolorbox}[breakable, colback=white, colframe=black!75, 
  fonttitle=\bfseries, title={Prompt Example: Agentic CoT - 2. Prediction module},
  boxrule=0.5pt, arc=2pt, outer arc=2pt]
\label{textbox:genagent_predictor}

\textbf{User}

[Participant's information]

Age: 50-64

Race or ethnicity: White

...

\bigskip
[Participant's prior responses]

Question: How much, if at all, do you think the following proposals would do to reduce economic inequality in the U.S.? Expanding government benefits for the poor

...

\bigskip
[Expert social scientist's observations/reflections]

(\textbf{Generated observations/reflections from the expert from step 1})

\bigskip
=====

\bigskip

What you see above is a participant information. Based on the information, I want you to predict the participant’s survey responses. All questions are multiple choice where you must guess from one of the options presented. As you answer, I want you to take the following steps:

Step 1) Describe in a few sentences the kind of person that would choose each of the response options. (``Option Interpretation")

Step 2) For each response option, reason about why the Participant might answer with the particular option. (``Option Choice")

Step 3) Write a few sentences reasoning on which of the option best predicts the participant’s response (``Reasoning")

Step 4) Predict how the participant will actually respond in the survey. Predict based on the information and your thoughts, but ultimately, DON’T overthink it. Use your system 1 (fast, intuitive) thinking. (``Response”)

\bigskip
Here is the question:

\bigskip
=====

\bigskip
Question: Thinking about the nation's economy,
how would you rate economic conditions in this country today?

A. Excellent

B. Good

C. Only fair

D. Poor

\bigskip
=====

\bigskip
Output format - output your response in json, where you provide the following:

\bigskip
\text{\{}``Response": ``$<$your predicted response option letter$>$"\text{\}}

\end{tcolorbox}

%% file: appendix_sections/note_gpt_oss.tex
\section{Tokenization: Qwen3 and GPT-OSS}
\label{appendix:note_gpt_oss}

In this section, we outline the differences in input preprocessing for GPT-OSS \citep{openai2025gptoss120bgptoss20bmodel} and Qwen-3 \citep{yang2025qwen3},
which arise from their distinct response formats.

\paragraph{GPT-OSS.}
GPT-OSS employs the Harmony response format to support advanced context engineering. 
Each generation typically begins with an analysis channel `\text{<|channel|>analysis}', where the model produces an internal chain-of-thought not exposed to users,
and concludes with a final channel `\text{<|start|>assistant<|channel|>final<|message|>}', which contains the user-facing response. 

During baseline experiments before fine-tuning,
to measure the model’s existing predictive capability, we place no constraints on generation:
the model is free to produce both analysis and final content,
and we parse the output from the final channel to evaluate accuracy.

During fine-tuning, however, we constrain the output to directly generate the answer in the final channel.
This step improves predictive accuracy while avoiding social bias that could result from fine-tuning on model-generated chain-of-thoughts, which may yield correct answers through ungrounded reasoning about individuals.
In this setup, we append the channel header explicitly to indicate the model that final answer should be generated, and apply next-token prediction loss to the final answer token.
An example training input prompt is shown below:

\begin{lstlisting}
<|start|>developer<|message|># Instructions

Respond to the following question by choosing one of the available options, and strictly answering with the option letter (e.g., 'A', 'B', etc.). Do not provide any additional text or explanation.

<|end|><|start|>user<|message|>Answer the following question as if your personal information is as follows:

Personal identification number: 12345.0
Age: 50-64
Race or ethnicity: White
Gender: Female
Education level: Some college, no degree
Income level: less than $30,000
Region of residence: West
Religion: Nothing in particular
Political party affiliation: Independent
Political ideology: Moderate

Question: Would you say the following was a reason or was not a reason why  there were guns in your household when you were growing up? For protection

A. Yes, was a reason
B. No, was not a reason

Answer:<|end|><|start|>assistant<|channel|>final<|message|>
\end{lstlisting}

As shown on the example, the tokenization step involves appending special tokens indicating the final channel.
Given the input prompt, the model generates probability distribution over available options in the next token.
Cross entropy loss is applied at that token position to fine-tune the model.

\paragraph{Qwen-3.}
Similarly, Qwen-3 introduces a thinking mode designed to let the model do more step-by-step reasoning (chain-of-thought) before generating a final answer.
During baseline experiments before fine-tuning, we place no constraints on generation and this allows the model to perform thinking
(wrapped by \texttt{<think>...</think>}).
During fine-tuning, we constrain the output to directly generate the answer by
appending the empty thinking (\texttt{<think>\textbackslash n\textbackslash n</think>}) explicitly to indicate the model for direct answer generation.